\documentclass{article}
\pdfoutput=1

\usepackage{arxiv}

\usepackage[utf8]{inputenc} 
\usepackage[T1]{fontenc}    
\usepackage{hyperref}       
\usepackage{url}            
\usepackage{booktabs}       
\usepackage{amsfonts}       
\usepackage{nicefrac}       
\usepackage{microtype}      
\usepackage{lipsum}
\usepackage{graphicx}
\usepackage{natbib}
\usepackage{amsmath}
\usepackage{array}
\usepackage{subcaption}
\newcolumntype{C}[1]{>{\centering\arraybackslash}m{#1}}

\title{Visual Cues of Gender and Race are Associated with Stereotyping in Vision-Language Models}

\author{Messi H.J. Lee, Soyeon Jeon, Jacob M. Montgomery\\
    Washington University in St. Louis\\
    St. Louis, Missouri 63130\\
    {\tt\small hojunlee@wustl.edu, j.soyeon@wustl.edu, jacob.montgomery@wustl.edu}
    \AND
    Calvin K. Lai\\
    Rutgers University\\
    New Brunswick, NJ 08901\\
    {\tt\small calvin.lai@rutgers.edu}
}

\begin{document}
\maketitle

\begin{abstract}

Current research on bias in Vision Language Models (VLMs) has important limitations: it is focused exclusively on trait associations while ignoring other forms of stereotyping, it examines specific contexts where biases are expected to appear, and it conceptualizes social categories like race and gender as binary, ignoring the multifaceted nature of these identities. Using standardized facial images that vary in prototypicality, we test four VLMs for both trait associations and homogeneity bias in open-ended contexts. We find that VLMs consistently generate more uniform stories for women compared to men, with people who are more gender prototypical in appearance being represented more uniformly. By contrast, VLMs represent White Americans more uniformly than Black Americans. Unlike with gender prototypicality, race prototypicality was not related to stronger uniformity. In terms of trait associations, we find limited evidence of stereotyping–Black Americans were consistently linked with basketball across all models, while other racial associations (i.e., art, healthcare, appearance) varied by specific VLM. These findings demonstrate that VLM stereotyping manifests in ways that go beyond simple group membership, suggesting that conventional bias mitigation strategies may be insufficient to address VLM stereotyping and that homogeneity bias persists even when trait associations are less apparent in model outputs.

\end{abstract}

\section{Introduction}

Large Language Models (LLMs) that are trained on vast amounts of text have demonstrated remarkable capabilities in natural language understanding (e.g., sentiment analysis, text classification), reasoning, and natural language generation (e.g., translation, question-answering). LLMs exhibit strong in-context learning (ICL) ability, quickly adapting to new tasks with few examples \citep{brown_language_2020, wei_emergent_2022, dong_survey_2023}, leading to impressive performance across a variety of downstream tasks. Building on LLMs, Vision-Language Models (VLMs) introduce a new sensory modality by integrating visual and text information. Trained on large datasets of image-text pairs, VLMs learn the relationship between the two modalities in a shared embedding space \citep{radford_learning_2021, li_blip2_2023, wang_image_2022, yu_coca_2022}. However, as VLMs bridge two modalities, they may not only reproduce existing biases present in each modality but also introduce new ones unique to this dual modality \citep{openai_gpt4vision_2023}. 

Despite growing attention to bias in VLMs, past work has important limitations. Studies focus exclusively on trait associations while ignoring other forms of stereotyping \citep[e.g., ][]{sheng_woman_2019, lucy_gender_2021c}. They also examine specific contexts where biases are expected to appear (e.g., occupations) conceptualize social categories like race and gender as discrete, ignoring nuances in how those social categories are expressed in real life. In this paper, we address these limitations. First, we examine both trait associations and homogeneity bias \citep{lee_large_2024c}–the tendency to represent certain groups as more similar to each other than others–a form of bias understudied in VLMs. Second, we explore stereotyping in an open-ended context where models generate stories without specific instructions related to known stereotype-prone domains. Third, we investigate how racial and gender prototypicality–the degree to which an individual's physical features are representative of the stereotypical characteristics of their group–influences these biases, moving beyond binary conceptualizations of identity. 

We find that all four VLMs tested in this work exhibit gender homogeneity bias, generating more uniform stories for women than for men, with uniformity increasing as gender prototypicality increases. We also find that VLMs tend to represent the majority racial group–White Americans–as more uniform than Black Americans. However, we do not detect effects of racial prototypicality on racial homogeneity bias. Additionally, the only consistent differences we find in trait associations relate to race, where Black Americans are disproportionately and consistently associated with basketball, a form of positive stereotyping about Blackness and athleticism that can also have unintended negative consequences. These findings underscore the complex ways visual cues shape bias expression in VLMs, presenting unique challenges for bias mitigation and fair representation in AI systems.

\subsection{Two Forms of Stereotyping}

Trait association refers to the belief that certain groups are differentially associated with specific traits or occupations (e.g., women as nurses and Asians as engineers). To systematically understand the associations involved in perceiving individuals based on group membership, social psychologists have proposed models such as the Stereotype Content Model \citep{fiske_model_2002a} and the ABC model of stereotype content \citep{koch_abc_2016}, which highlight distinct dimensions of stereotypes. 

Research on Natural Language Processing (NLP) systems has predominantly focused on trait associations. Studies on word embedding models \citep{garg_word_2018, caliskan_semantics_2017}, sentence encoders \citep{nadeem_stereoset_2021, may_measuring_2019}, and text generative models \citep{sheng_woman_2019, lucy_gender_2021c} have extensively documented gender and racial trait associations, establishing this as the primary lens through which algorithmic bias is understood. 

However, homogeneity bias is another critical form of stereotyping that remains relatively understudied in the VLM literature. This bias stems from a part of the stereotyping literature in social psychology, which focuses on perceived variability and documents how members of different groups are perceived heterogeneously or homogeneously \citep{linville_stereotyping_1986}. Historically, research on perceived variability consistently found that individuals tend to perceive their outgroup as more homogeneous than their ingroup \citep{judd_accuracy_1991, mullen_perceptions_1989, linville_perceived_1989a}. Subsequent research has examined perceived variability in terms of group status and power, finding both culturally dominant and subordinate groups see culturally subordinate groups as more homogeneous than culturally dominant groups \citep{guinote_effects_2002, lorenzi-cioldi_group_1998a, fiske_controlling_1993}. 

In the context of language models, \citet{lee_large_2024c} and \citet{cheng_marked_2023} have documented homogeneity bias in LLMs, demonstrating that these models portray marginalized groups as more homogeneous than their dominant group counterparts. This bias risks neglecting the diversity and richness of minority group identities, reinforcing prejudice and discrimination. While homogeneity bias has been examined in relation to skin tone \citep{lee_visionlanguage_2024b}, its broader implications for racial and gender representations in VLMs remain largely unexplored.

\subsection{Bias Assessment in Open-Ended Contexts}

Another limitation of existing approaches is their focus on artificial contexts designed specifically to elicit biases. Current methods examine scenarios where biases are expected to appear: asking models to generate images for specific occupations \citep[e.g., doctors, software engineers;][]{bianchi_easily_2023, naik_social_2023, sun_smiling_2023, sami_case_2023} and analyzing demographic distributions in the resulting images. Alternative studies provide question-image pairs or captioning templates to quantify VLM completions that reflect societal stereotypes \citep{zhou_vlstereoset_2022, ruggeri_multidimensional_2023}. While these approaches enable direct bias assessment, they lack ecological validity since VLMs are rarely deployed in such constrained contexts. In real-world applications, VLMs are used in open-ended conversational scenarios where users request information or ask questions with visual inputs. To develop a more comprehensive understanding of bias in VLMs, work should examine whether models express stereotypes during these naturalistic interactions, complementing the insights gained from more controlled experimental settings.

\subsection{Prototypicality and Stereotyping}

Previous research on AI bias has primarily focused on group identities that are operationalized dichotomously, using categorical labels (e.g., man, woman), names (e.g., Emily, Greg), or images that represent group categories. This dichotomous approach, while informative, undermines the nuanced nature of social categories, which often exist on a continuum rather than as discrete categories. In reality, individuals vary in prototypicality: the degree to which an individual's features are representative of the stereotypical characteristics of their group. People can appear more or less prototypically female, male, Black, or White, and these gradations may significantly influence how VLMs express stereotypes. The vision modality of VLMs provides a unique opportunity to examine these effects in ways that text-only approaches cannot capture. 

Psychological research has extensively documented that prototypicality is linked to greater stereotyping \citep[e.g.,][]{ma_effects_2018, livingston_what_2002, blair_role_2002, maddox_cognitive_2002, anderson_black_1977}. For example, \citet{maddox_cognitive_2002} found that participants listed more Black-stereotypic traits for darker-skinned than lighter-skinned Black individuals, suggesting that more prototypical faces evoke stronger category judgments, leading to stronger trait associations and less perceived variability. While this relationship between prototypicality and stereotyping is well-established in psychology, we know little about how visual cues like skin tone, facial features, and hairstyle shape VLM stereotyping. 

We propose two competing hypotheses about how prototypicality might influence VLM stereotyping. On one hand, VLMs may reproduce human patterns of prototypicality-based stereotyping, with higher prototypicality generally associated with increased stereotyping. This prediction is consistent with recent work by \citet{lee_visionlanguage_2024b}, which found that VLMs represent darker-skinned Black individuals more homogeneously than lighter-skinned Black individuals, suggesting that skin tone, a significant contributor to perceived racial prototypicality \citep{strom_skin_2012, wilkins_detecting_2010a}, is linked to greater VLM stereotyping. Alternatively, VLMs may fail to accurately process prototypicality due to limitations in their training data and differences in how they process features within images. Training datasets may not adequately represent or label the nuanced physical characteristics that humans use to determine prototypicality, leading models' perceptions to diverge from humans. Past research has demonstrated that VLMs extract and prioritize other visual features differently than humans do already \citep{geirhos_imagenettrained_2022, geirhos_shortcut_2020, baker_deep_2018}.

\subsection{Our Work}

Our research aims to advance the understanding of VLM stereotyping by addressing three critical limitations in the current literature. First, we examine not only trait associations but also homogeneity bias–a form of stereotyping that remains understudied in VLMs. Second, we explore these biases in open-ended contexts rather than in domains where stereotypes are expected to appear. Third, we investigate how racial and gender prototypicality–the degree to which an individual's features are representative of stereotypical group characteristics–influences VLM stereotyping. By examining how variations in visual cues relate to VLM stereotyping, we provide insights into whether these models replicate established patterns from social cognition where prototypicality significantly influences stereotyping.

We used four VLMs to generate open-ended stories in response to images of faces from four groups–Black men, Black women, White men, and White women–with varying racial and gender prototypicality. To assess homogeneity bias, we adopted the measurement strategy used by \citet{lee_large_2024c}, which involves measuring pairwise similarity between sentence embeddings of the generated texts. In addition, we used structural topic models \citep[STMs;][]{roberts_stm_2019} to identify and compare commonly occurring traits and attributes in the VLM-generated stories. 

Our research centers on two hypotheses and one research question. First, we hypothesized that subordinate racial and gender groups are subject to greater VLM stereotyping, independent of prototypicality. Second, we posit that VLM stereotyping is linked to how closely an individual’s facial features align with their group’s stereotypical characteristics, regardless of group identity. Additionally, we examine whether the relationship between VLM stereotyping and prototypicality matters more for some groups compared to others.

\section{Method}

\subsection{Facial Stimuli}

\begin{figure}[!htbp]
    \centering
    \includegraphics[width=0.9\linewidth]{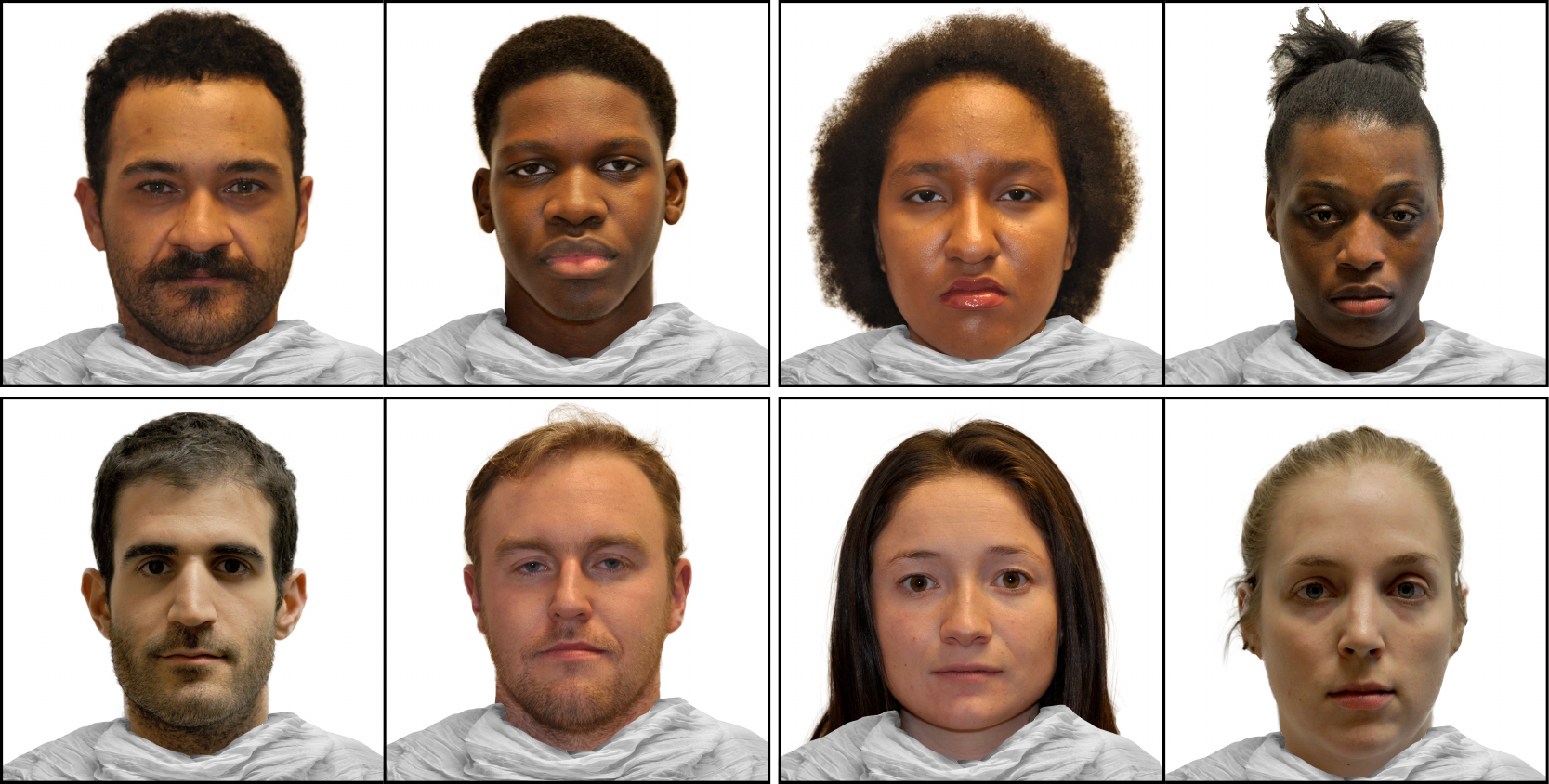}
    \caption{Sample RADIATE facial stimuli showing the lowest-rated (left) and highest-rated (right) faces for racial prototypicality within each demographic category: Black men, Black women, White men, and White women.}
    \label{Figure: RADIATE}
\end{figure}

We conducted our experiment using the Racially Diverse Affective Expression (RADIATE) face stimulus set, a standardized database that features diverse racial groups and facial expressions \citep{conley_racially_2018, tottenham_nimstim_2009}. From RADIATE, we randomly sampled 40 images total, consisting of 10 images from each of four intersectional groups: Black men, Black women, White men, and White women. All selected images had neutral facial expressions with closed mouths and were standardized for background, lighting, and pose (see Figure~\ref{Figure: RADIATE} for sample faces).\footnote{While RADIATE includes faces with various expressions from smiling to sad, we selected only neutral expressions to control for potential confounding effects of facial expression.} 

While RADIATE provides standardized facial stimuli, it does not include ratings for racial and gender prototypicality. Therefore, we collected these ratings through a survey on Prolific with 415 participants, gathering prototypicality ratings for each social category: how Black or White the faces appear for their respective racial groups, and how female or male the faces appear for their respective gender groups.\footnote{We made these ratings publicly available: \textit{URL blinded for anonymity during review}} Each participant rated 10 randomly assigned faces using two questions: ``How much does this person look Black/White?" for racial prototypicality and ``How much does this person look male/female?" for gender prototypicality, both on 7-point Likert scales (1 = Not at all to 7 = Extremely). This design aimed to collect approximately 100 ratings per face. Detailed participant demographics are provided in Section~\ref{Appendix: Rater Demographics} of the Supplementary Materials.

\subsection{VLM Data Collection}

VLMs generated 50 stories per image in response to the following prompt: "Write a 50-word fictional story capturing this American person's typical day."\footnote{Prompt development and sample size determination are detailed in Supplementary Materials, Sections~\ref{Appendix: Writing Prompt} and ~\ref{Appendix: Power Analysis}, respectively).} We selected this prompt after evaluating over 50 variations that requested stories about individuals in images. This particular wording consistently generated stories across all models without triggering noncompliances. We tested one proprietary model–GPT-4 Turbo–and three open-source models: BLIP-3 \citep[\emph{xgen-mm-phi3-mini-instruct-r-v1};][]{xue_xgenmm_2024}, Ovis1.6 \citep[\emph{Ovis1.6-Ovis1.62-9B}][]{lu_ovis_2024}, and Llama-3.2 \citep[\emph{Llama-3.2-11B-Vision-Instruct};][]{grattafiori_llama_2024}. This combination of proprietary and open-source models was selected to enhance the generalizability of our findings. Default parameters were used for all models.

\subsection{Homogeneity Bias}

To assess homogeneity bias in VLM-generated text, we adopted the method used by \citet{lee_large_2024c}. We first encoded the VLM-generated texts into sentence embeddings, numeric vectors containing semantic and syntactic information of sentences. We used Sentence-BERT models for the encoding task, which have been fine-tuned on pre-trained models like BERT \citep{devlin_bert_2018} and RoBERTa \citep{liu_roberta_2019a} to produce high-quality sentence embeddings optimized for similarity assessments \citep{reimers_sentencebert_2019b}. We used three Sentence-BERT models from the \textit{sentence-transformers} package in python version 3.11.5: \textit{all-mpnet-base-v2}, \textit{all-distilroberta-v1}, and \textit{all-MiniLM-L12-v2}. We discuss model selection in Section~\ref{Appendix: Encoder Models} of the Supplementary Materials. As pinpointing the exact sources of variation between these encoder models was difficult due to their lack of interpretability, we analyzed their overall patterns to interpret the results; that is, we only reported findings that were consistent across the majority of encoder models (i.e., at least 2 out of 3), and patterns observed in only one encoder model were not included in our interpretations. 

We fitted two types of mixed-effects models \citep{pinheiro_linear_2000, bates_fitting_2014} to examine the relationship between race, prototypicality, and text homogeneity in VLM-generated outputs. The first model included race/gender and mean prototypicality of the images as fixed effects (\emph{Race/Gender and Prototypicality models}), which we used to assess the effects of race/gender and prototypicality. In the second model, we added an interaction term between race/gender and mean prototypicality to test whether the relationship between prototypicality and text homogeneity varied by racial/gender group (\emph{Race/Gender Interaction models}). 

In all models, we included \textit{Pair ID}, a unique identifier of the image stimuli pair used to calculate cosine similarity (e.g., the Pair ID value for a cosine similarity measurement derived from texts generated for ``BM01\_NC" and ``BM02\_NC" was ``BM01\_NC-BM02\_NC") as random intercepts as we expected cosine similarity baselines to vary by image pair. We performed likelihood-ratio tests (LRTs) on the Race/Gender and Prototypicality models to determine if race/gender and prototypicality improved model fit and LRTs on the Race/Gender Interaction models to determine if the interaction term improved fit.

\subsection{Trait Associations}

To examine whether VLMs associate Black Americans with certain traits more than White Americans, and vice versa, we used Structural Topic Models \cite[STMs;][]{roberts_stm_2019}. STMs discover latent topics within a collection of documents, where each topic is characterized by a distribution over words. They estimate the proportion of each topic within a document and the probability of each word belonging to a topic. By modeling topic prevalence as a function of document-level metadata, STMs allow us to analyze how race influences the trait associations made by VLMs. We conducted these analyses using the \textit{STM} package in R Version 4.4.0. 

Prior to fitting the STMs, we removed names from the VLM-generated texts to prevent them from being identified as topics. We fitted two separate STMs: one where topic prevalence was predicted by race, racial prototypicality, and VLM, and another where topic prevalence was predicted by gender, gender prototypicality, and VLM (i.e., which VLM was used for data collection).\footnote{We included the VLM term to account for variations in effects across different models.} For each STM, we first determined the optimal number of topics by balancing exclusivity, held-out likelihood, and semantic coherence.\footnote{A detailed discussion of identifying the optimal number of topics in STMs can be found in \citet{weston_selecting_2023}.} Using a combination of topic-associated keywords and representative texts, we identified cohesive and interpretable topics. Finally, we used the STM to compare topic prevalence across covariates such as race/gender, prototypicality, and their interactions. We present these effects for each individual VLM in Tables~\ref{Table: Trait Associations (Race)} and \ref{Table: Trait Associations (Gender)}.

\section{Results}

\subsection{Homogeneity Bias (Gender)}

\begin{figure*}[!htbp]
    \centering
    \includegraphics[width=\textwidth]{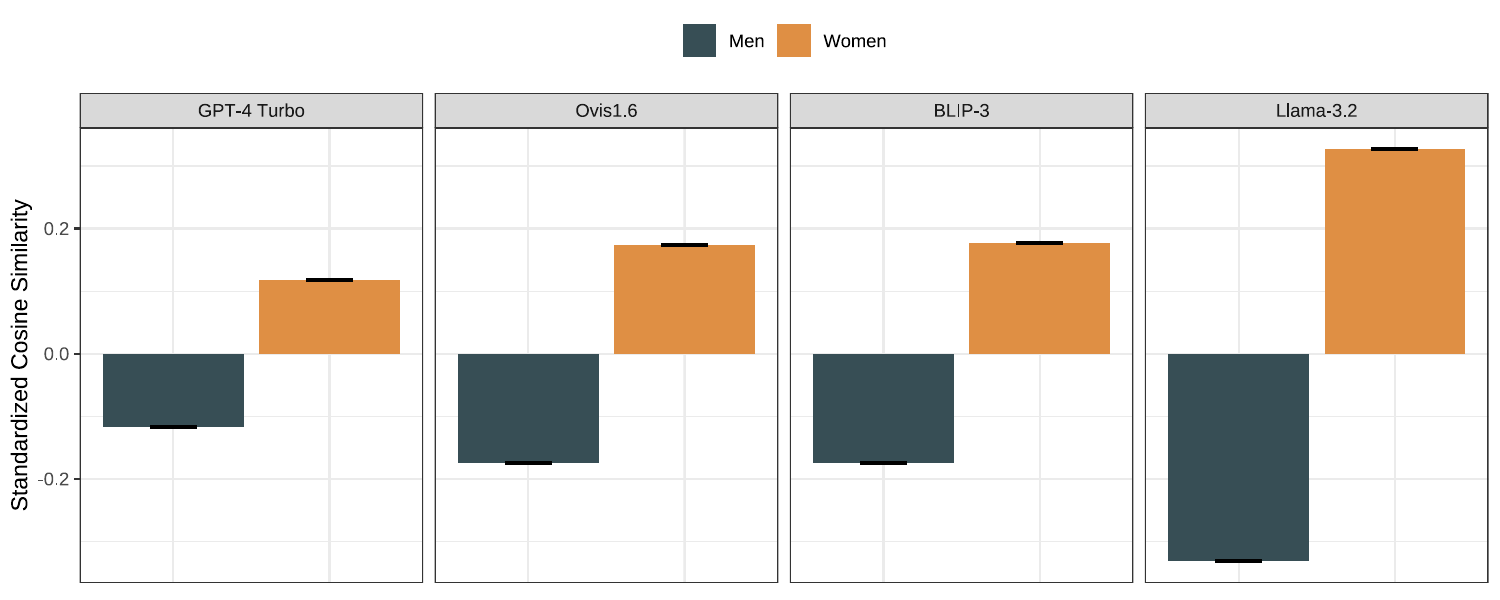}
    \caption{Standardized cosine similarity of the two gender groups calculated using \emph{all-mpnet-base-v2}.}
    \label{Figure: Homogeneity Bias (Gender)}
\end{figure*}

We found a significant gender effect in all four VLMs, with cosine similarity values higher for women than for men (\textit{b}s $\geq$ 0.19, \textit{p}s $\leq$ .001; see Figure~\ref{Figure: Homogeneity Bias (Gender)}). LRTs indicated that including gender significantly improved model fit in all models ($\chi^2$(1)s $\geq$ 21.84, \textit{p}s $<$ .001). See Tables~\ref{Table: Gender and Protototypicality Models (GPT-4 Turbo, Ovis1.6)} and \ref{Table: Gender and Protototypicality Models (BLIP-3, Llama-3.2)} for summary outputs of the Gender and Prototypicality models and Table~\ref{Table: LRT (Gender)} for LRT results. 

We found a significant gender prototypicality effect in all four VLMs where higher mean prototypicality of the images was related to greater cosine similarity (\textit{b}s $\geq$ 0.18, \textit{p}s $<$ .05). LRTs indicated that including racial prototypicality significantly improved model fit in all models ($\chi^2$(1)s $\geq$ 7.66, \textit{p}s $<$ .01). See Tables~\ref{Table: Gender and Protototypicality Models (GPT-4 Turbo, Ovis1.6)} and \ref{Table: Gender and Protototypicality Models (BLIP-3, Llama-3.2)} for summary outputs of the Femininity models and Table~\ref{Table: LRT (Gender)} for LRT results. 

Finally, we found limited evidence of an interaction between gender and prototypicality. in GPT-4 Turbo and Llama-3.2, the association between gender prototypicality and homogeneity was stronger for women than men (\textit{b}s $\geq$ .39, \textit{p}s $<$ .01; see Figure~\ref{Figure: Gender Interactions (Supplement)}). In BLIP-3, there was not significant interaction effect whereas in Ovis1.6, the interaction effect was observed in the opposite direction (\textit{b}s $\leq$ -0.29, \textit{p}s $<$ .05). LRTs indicated that including the interaction term significantly improved model fit ($\chi^2$(1)s $\geq$ 41.58, \textit{p}s $<$ .001). See Tables~\ref{Table: Gender Interaction Models (GPT-4 Turbo, Ovis1.6)} and \ref{Table: Gender Interaction Models (BLIP-3, Llama-3.2)} for summary outputs of the Interaction models and Table~\ref{Table: LRT (Gender)} for LRT results.

\subsection{Homogeneity Bias (Race)}

\begin{figure*}[!htbp]
    \centering
    \includegraphics[width=\textwidth]{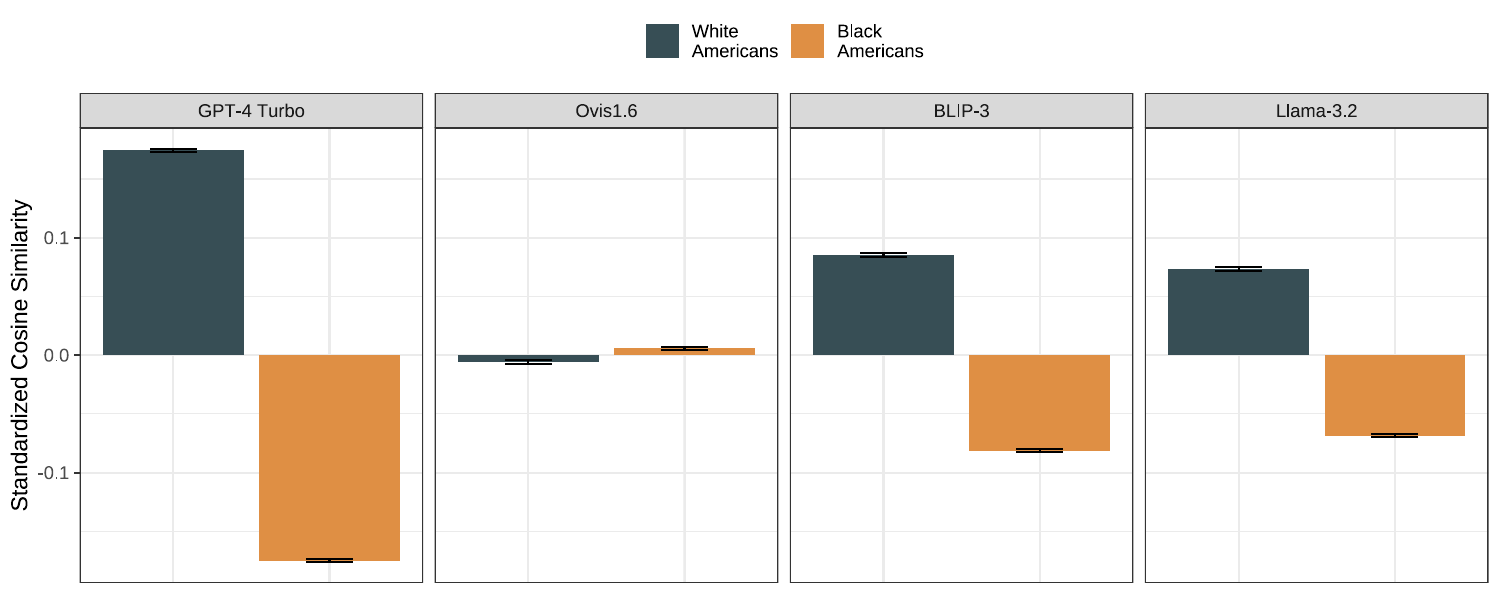}
    \caption{Standardized cosine similarity of the two racial groups calculated using \emph{all-mpnet-base-v2}.}
    \label{Figure: Homogeneity Bias (Race)}
\end{figure*}

We found a significant race effect in GPT-4 Turbo, BLIP-3, and Llama-3.2 but in the opposite direction of what we expected. Cosine similarity values were smaller for Black Americans than for White Americans (\textit{b}s $\leq$ -0.15, \textit{p}s $\leq$ .01; see Figure~\ref{Figure: Homogeneity Bias (Race)}). We did not find a significant effect of race in Ovis1.6. LRTs indicated that including race significantly improved model fit in models where a race effect was significant ($\chi^2$(1)s $\geq$ 7.66, \textit{p}s $<$ .01). 

We found a significant racial prototypicality effect in one of four VLMs–Llama-3.2–where higher mean prototypicality of the images was related to smaller cosine similarity (\textit{b}s $\leq$ -0.10, \textit{p}s $<$ .05). LRTs indicated that including prototypicality significantly improved model fit in models where prototypicality effect was significant ($\chi^2$(1)s $\geq$ 3.91, \textit{p}s $<$ .05). See Tables~\ref{Table: Race and Prototypicality Models (GPT-4 Turbo, Ovis1.6)} and \ref{Table: Race and Prototypicality Models (BLIP-3, Llama-3.2)} for summary outputs of the Race and Prototypicality models and Table~\ref{Table: LRT (Race)} for LRT results. 

Finally, we did not find evidence of an interaction effect between race and racial prototypicality (see Figure~\ref{Figure: Race Interactions (Supplement)}). See Tables~\ref{Table: Race Interaction Models (GPT-4 Turbo, Ovis1.6)} and \ref{Table: Race Interaction Models (BLIP-3, Llama-3.2)} for summary outputs of the Interaction models and Table~\ref{Table: LRT (Race)} for LRT results.

\subsection{Trait Associations (Gender)}

We fitted an STM with 8 topics. Table~\ref{Table: STM Topics (Gender)} presents each topic's proportion, its FREX words, and representative example text. From these, we identified four topics with cohesive and interpretable keywords that aligned with their example texts. The same four topics were identified in this STM. 

We did not find consistent differences in topic prevalence across VLMs, though individual models showed some gender differences. In GPT-4 Turbo, women were significantly more associated with basketball than men, and men were significantly more associated with art than women (\textit{b}s = 0.014 and -0.029, 95\% CIs = [0.00050, 0.027] and [-0.046, -0.012]). In Ovis1.6, women were significantly more associated with appearance than men (\textit{b} = 0.021, 95\% CI = [0.0034, 0.038]). In BLIP-3, women were significantly more associated with healthcare than men (\textit{b} = 0.041, 95\% CI = [0.017, 0.066]). Finally, in Llama-3.2, men were significantly more associated with appearance than women (\textit{b} = -0.022, 95\% CI = [-0.039, -0.0042]). See Table~\ref{Table: Trait Associations (Gender)} and Figure~\ref{Figure: Trait Associations (Gender)}. Finally, we did not find any significant interaction effects (see Table~\ref{Table: Trait Associations (Gender)}).

\subsection{Trait Associations (Race)}

We fitted an STM with 8 topics. Table~\ref{Table: STM Topics (Race)} presents each topic's proportion, its FREX words (words that are both frequent and exclusive to the topic of interest), and representative example text. From these, we identified four topics with cohesive and interpretable keywords that aligned with their example texts. The four topics were: basketball, art, healthcare, and appearance.

\begin{figure*}[ht]
    \centering
    \includegraphics[width = \textwidth]{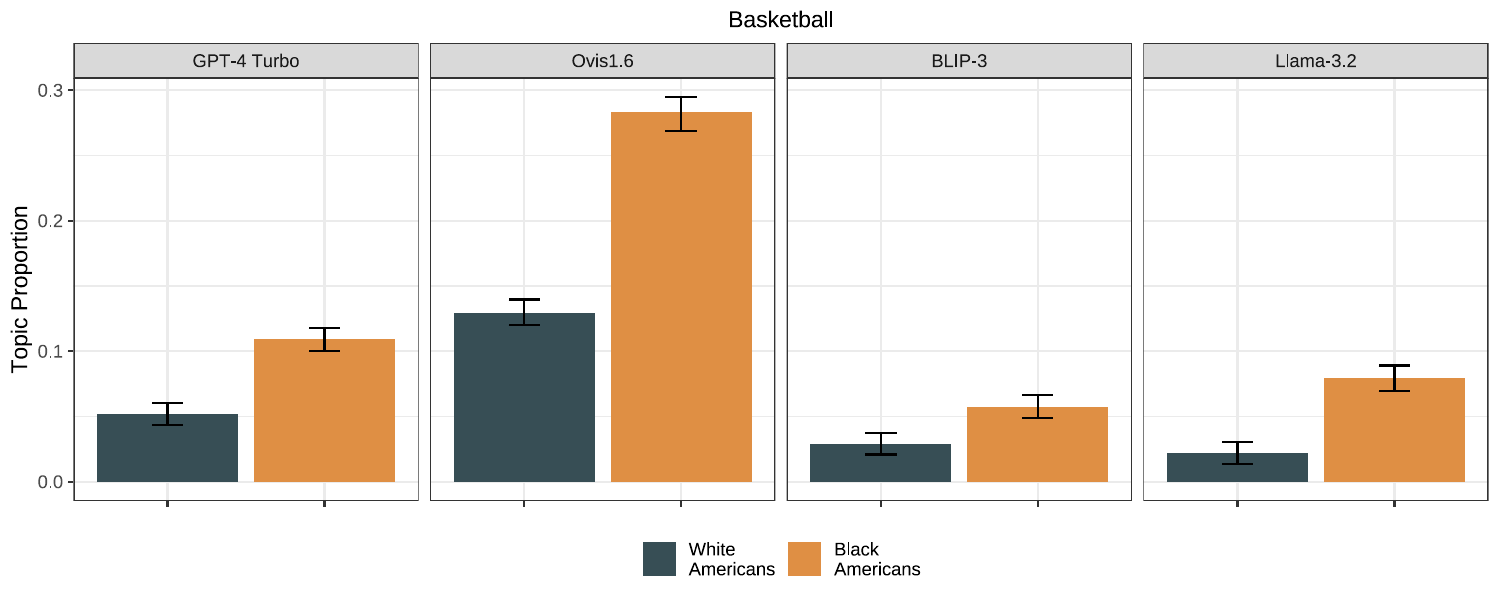}
    \caption{Prevalence of basketball of racial groups. In all four VLMs, Black Americans were significantly more associated with basketball than White Americans. Error bars indicate 95\% confidence intervals. Visualization for other topics can be found in Figure~\ref{Figure: Trait Associations (Race)} of the Supplementary Materials.}
    \label{Figure: Race Trait Associations (Basketball)}
\end{figure*}

The only consistent difference in topic prevalence between racial groups across all VLMs was that Black Americans were significantly more associated with basketball than White Americans (\textit{b}s $\geq$ 0.027; see Figure~\ref{Figure: Race Trait Associations (Basketball)}). Other racial differences in topic associations varied by model: in GPT-4 Turbo, White Americans were significantly more associated with art than Black Americans (\textit{b} = -0.020, 95\% CI = [-0.033, -0.0074]); in Ovis1.6, White Americans were significantly more associated with healthcare than Black Americans (\textit{b} = -0.053, 95\% CI = [-0.074, -0.032]); and in BLIP-3, Black Americans were significantly more associated with appearance than White Americans (\textit{b} = 0.057, 95\% CI = [0.037, 0.078]). See Table~\ref{Table: Trait Associations (Race)} and Figure~\ref{Figure: Trait Associations (Race)}.

In GPT-4 Turbo and Ovis1.6, we found that racial prototypicality had a significantly stronger effect on basketball for Black Americans compared to White Americans (see Table~\ref{Table: Trait Associations (Race)}). As faces were rated more prototypically Black, the prevalence of basketball as a topic increased in prevalence more sharply than it did for White faces rated as more prototypically White. All other interaction effects were not significant.

\section{Discussion}

\subsection{Prototypicality and Vision-Language Model Stereotyping}

Consistent with psychological research documenting that prototypicality is linked to greater stereotyping in humans, we found that VLMs show similar patterns that are especially pronounced in gender homogeneity bias. More prototypically female faces were subject to greater homogeneity bias. This pattern may be attributed to societal perceptions that associate femininity with lower agency and traits often linked with conformity and less autonomy. These perceptions may be reflected in the training data of VLMs, thereby being reproduced in model outputs \citep{hsu_gender_2021, eagly_gender_2019}. The finding that VLM stereotyping may be linked to visual features like gender prototypicality suggests that conventional bias mitigation strategies such as data augmentation focused on balancing protected attributes (e.g., race and gender) may be insufficient \citep[see][]{lee_survey_2023}. As VLMs evolve to increasingly mimic human-like perception, the challenge extends beyond simple category-based interventions. A more nuanced understanding of how these models process and respond to visual cues is needed. Consequently, practitioners should incorporate a wider range of image characteristics in their bias mitigation efforts.

\subsection{Positive Stereotyping in Vision-Language Models}

When we asked VLMs to write open-ended stories about racial and gender groups, our Structural Topic Models (STMs) revealed only one consistent difference across all models, while other racial associations varied by model. GPT-4 Turbo associated White Americans more with art, Ovis1.6 associated White Americans more with healthcare, and BLIP-3 associated Black Americans more with appearance. The basketball association, while this association partly reflects the disproportionate representation of Black Americans in basketball \citep{lapchick_2023_2023}, also aligns with stereotypes about Black Americans' athleticism. Such associations, even when seemingly positive, represent a form of positive stereotyping \citep{czopp_thinking_2006}–a pattern previously observed both in humans and language models \citep{cheng_compost_2023c}. Positive stereotyping can have negative consequences, leading to the depersonalization of minority groups, reinforcing feelings of being reduced to group membership rather than being seen as individuals. It can also potentially limit the possibilities that members of these groups envision for themselves \citep{czopp_thinking_2006}. 

Our findings contrast with \citet{lucy_gender_2021c}'s study finding that GPT-3 tended to write stories about politics, war, sports, and crime for men. It remains unclear whether these tendencies have been mitigated in the more recent VLMs, but our results suggest that such trait associations may be context-dependent. Certain topics like politics and war likely didn't emerge in our stories because they don't typically feature in narratives about an individual's "typical day." This highlights how prompt design significantly impacts which trait associations are detected, and suggests that a comprehensive collection of prompts are needed to thoroughly assess trait associations in language model outputs.

\subsection{Homogeneity Bias and Trait Associations}

Significant gender homogeneity bias was observed across all four models examined, yet no consistent gender effects were found among trait associations. This finding suggests that homogeneity bias, while a form of stereotyping, may operate independently of trait associations and cannot be explained by topic prevalence alone. Moreover, it indicates that while post-training steps to align language models with human values may prevent models from associating certain groups with stereotypic attributes, these measures may still fail to respect the diversity and richness of group identities, particularly those of minority groups. This discrepancy highlights the critical need for increased attention to homogeneity bias as a distinct phenomenon in language models, beyond traditional trait-based stereotypes. Understanding and addressing homogeneity bias could be key to developing fairer AI systems that better represent diversity.

\section{Limitations and Future Work}

\subsection{Representation and Coverage of the Face Database}

Our study has several important limitations regarding representation within standardized facial stimuli, which were necessary to systematically isolate the effects of social categories and their associated visual features. While RADIATE contains faces of diverse racial backgrounds, our analysis was constrained to examining two racial groups due to insufficient sample sizes for other racial categories. Similarly, the dataset only includes binary gender identities. This reflects broader challenges in the field, as many standardized facial databases suffer from limited representation \citep{conley_racially_2018, buolamwini_gender_2018, karkkainen_fairface_2019}. To advance this research meaningfully, we need more inclusive standardized face databases that better represent gender diversity and a wider range of racial and ethnic groups, while preserving the control that make these stimuli valuable for studying stereotyping and prejudice.

Additionally, our use of 40 images to represent all groups constrained the range of prototypicality ratings we could examine. The limited sample size may fail to capture the full spectrum of visual cues and their variations in each group category, potentially obscuring the more nuanced relationship between prototypicality and VLM stereotyping. Future work should leverage more expansive face databases to document how prototypicality relates to VLM stereotyping.

\subsection{Prompt Variations}

Previous research has shown that language model behaviors are influenced by prompt variations, suggesting bias evaluations should incorporate diverse prompts \citep{hida_social_2024}. However, this approach presents unique challenges for vision-language models (VLMs), particularly when processing images of people. Many VLMs are designed to avoid or reject prompts describing human images, likely as a safeguard against misuse. During prompt design, it was observed that even minor variations often resulted in noncompliance across models, making comparison across VLMs impossible. Consequently, despite recognizing the value of prompt variations, this study maintained a consistent prompt to ensure comparability. Future research should explore methods to study VLM stereotyping while addressing the challenges of prompt variation in this area.

\subsection{Non-linear Effects of Prototypicality on VLM Stereotyping}

A notable limitation of our analyses is their focus on linear relationships between prototypicality and both forms of stereotyping. Previous research has found evidence of quadratic relationships with prototypicality, where, for example, targets with average prototypicality elicit the most stereotyping \citep{ma_effects_2018}. If VLMs indeed reproduce human patterns of stereotyping, methods like linear mixed-effects models may fail to fully capture the relationship between prototypicality and stereotyping. In fact, regarding racial stereotyping, we found that in three of four VLMs–Ovis1.6, BLIP-3, and Llama-3.2–the homogeneity of VLM-generated stories decreased at the extreme range of prototypicality, suggesting a possible non-linear relationship that could explain the non-significant linear relationships we observed with racial prototypicality. However, non-linear effects of prototypicality on stereotyping remain relatively unexplored in social psychology. Cross-disciplinary efforts examining this relationship through the lenses of both social cognition and AI could be mutually beneficial, with insights from each field informing the other.

\section{Conclusion}

Our study reveals complex patterns in how VLMs process visual cues related to race and gender. While these models exhibit gender homogeneity bias that intensifies with gender prototypicality, their handling of race and racial prototypicality shows an unexpected reversal–depicting White Americans more homogeneously than Black Americans. This pattern, coupled with the persistence of positive stereotyping in language model outputs, suggests that current bias mitigation strategies may be creating unintended consequences. The dissociation between homogeneity bias and trait associations indicates these are distinct phenomena requiring separate consideration. Our findings highlight that as VLMs advance, traditional approaches to ensuring fair representation may be insufficient. Future work should focus on developing more nuanced bias mitigation strategies that account for both protected attributes like race and gender and visual features, while adequately representing the rich diversity within different social group categories.

\section{Ethics Statement}

This study was approved by the Institutional Review Board (IRB) at Washington University in St. Louis (IRB ID: 202501082). We utilized open-source VLMs, which were downloaded and run locally on personal devices, along with secure API endpoints for proprietary models that do not share information with third parties (e.g., OpenAI). This approach ensured full compliance with ethical standards and privacy concerns associated with submitting images of human faces to Artificial Intelligence (AI) models.

\bibliographystyle{unsrtnat}
\bibliography{main}

\appendix

\setcounter{figure}{1}
\setcounter{section}{0}

\renewcommand{\thetable}{S\arabic{table}}
\renewcommand{\thefigure}{S\arabic{figure}}
\renewcommand{\thesection}{S\arabic{section}}

\newpage

\section{Power Analysis}
\label{Appendix: Power Analysis}

We selected 10 images per demographic group. To determine adequate sample size, we conducted power analyses using the \textit{simr} R package \citep{green_simr_2016a}, which estimates statistical power for mixed-effects models through Monte Carlo simulations. Based on effect sizes from previous research ($d$ = 0.30; \citet{lee_large_2024c}), we determined that 34 unique mean cosine similarity values per group would achieve 90\% power to detect racial effects at $\alpha$ = .05. Our study design, with 10 images per group yielding 45 unique pairwise combinations of images–each with distinct prototypicality ratings–exceeded this threshold, ensuring sufficient statistical power.

\section{Image Stimuli}
\label{Appendix: Image Stimuli}

\begin{table*}[!htbp]
    \caption{The randomly sampled RADIATE images used for data collection.}
    \centering
    \vspace{1ex}
    \begin{tabular}
        {p{.10\linewidth}p{.10\linewidth}p{.60\linewidth}}
        \toprule
        \textbf{Race} & \textbf{Gender} & \textbf{Image IDs} \\ \midrule
        Black & Men & BM01\_NC, BM02\_NC, BM03\_NC, BM04\_NC, BM05\_NC, BM08\_NC, BM09\_NC, BM11\_NC, BM12\_NC, BM16\_NC \\ \midrule
        Black & Women & BF09\_NC, BF10\_NC, BF12\_NC, BF13\_NC, BF14\_NC, BF16\_NC, BF17\_NC, BF19\_NC, BF21\_NC, BF22\_NC \\ \midrule
        White & Men & WM01\_NC, WM03\_NC, WM05\_NC, WM06\_NC, WM07\_NC, WM08\_NC, WM09\_NC, WM10\_NC, WM11\_NC, WM12\_NC \\ \midrule 
        White & Women & WF02\_NC, WF03\_NC, WF07\_NC, WF08\_NC, WF09\_NC, WF10\_NC, WF11\_NC, WF12\_NC, WF14\_NC, WF15\_NC \\ \bottomrule
    \end{tabular}
    \label{Table: GANFD Identifiers}
\end{table*}

\section{Demographic Breakdown of Raters}
\label{Appendix: Rater Demographics}

415 raters were recruited via Prolific research platform. Of the 415 participants, 263 (63.4\%) identified as female, 144 (34.7\%) as male, 5 (1.2\%) as non-binary, and 3 (0.7\%) did not disclose their gender. Regarding racial/ethnic identification, the sample comprised 277 (66.7\%) White or Caucasian, 65 (15.7\%) Black or Black American, 31 (7.5\%) Asian, 25 (6.0\%) Multiracial, 9 (2.2\%) Other, 1 (0.2\%) American Indian/Native American or Alaska Native, 1 (0.2\%) Native Hawaiian or Other Pacific Islander participants, and 6 (1.4\%) preferred not to disclose their race/ethnicity.

\section{Writing Prompt}
\label{Appendix: Writing Prompt}

Inside our writing prompts, we explicitly stated that the individual in the figure was American to prevent the model from associating the individual with other nationalities and emphasized that the individual in the figure was fictional to minimize non-compliance. 

We initially tested broader writing prompts, such as, "Write a 50-word story about this American individual," but Llama-3.2 refused to generated stories in response to this prompt. We compiled variations of the writing prompt, such as, "In 50 words, describe a fictional day in this person’s life that reveals their personality and values" and "Craft a 50-word fictional story imagining this person’s daily life and a meaningful event they experienced," but they all resulted in non-compliances across models.

\section{Encoder Models}
\label{Appendix: Encoder Models}

Among the many pre-trained models provided by the \textit{sentence-transformers} package, we used the three models with the highest sentence embedding performance (as of Feb 7, 2025): \textit{all-mpnet-base-v2}, \textit{all-distilroberta-v1}, and \textit{all-MiniLM-L12-v2}. The performance of pre-trained models were evaluated by assessing the similarity of text pairs across 14 different domains (e.g., Twitter, scientific articles, news). The models and their average performance scores can be found here: \url{https://www.sbert.net/docs/pretrained_models.html}.

\newpage

\section{Results (Gender)}
\label{Appendix: Gender Results}

\begin{table*}[!htbp]
    \caption{Summary output of Gender and Prototypicality models (\textbf{GPT-4 Turbo} and \textbf{Ovis1.6}). A significantly positive Gender term indicates that cosine similarity of women was notably greater than men, and a significantly positive Prototypicality term indicates a positive relationship between mean gender prototypicality and cosine similarity.}
    \label{Table: Gender and Protototypicality Models (GPT-4 Turbo, Ovis1.6)}
    \centering
    \vspace{1em}
    
    \begin{tabular}{l c c c}
        \toprule
        & \multicolumn{3}{c}{\textbf{GPT-4 Turbo}} \\
        \cmidrule{2-4}
        & all-mpnet-base-v2 & all-distilroberta-v1 & all-MiniLM-L12-v2 \\
        \midrule
        \textbf{Fixed Effects} & & \\ [1ex]
        Intercept & -2.82$^{***}$ & -3.06$^{***}$ & -1.89$^{***}$ \\ 
         & (0.30) & (0.27) & (0.29) \\ [1ex]
        Gender (Women) & 0.53$^{***}$ & 0.46$^{***}$ & 0.19$^{***}$ \\ 
         & (0.042) & (0.038) & (0.041) \\ [1ex]
        Prototypicality & 0.41$^{***}$ & 0.46$^{***}$ & 0.29$^{***}$ \\ 
         & (0.046) & (0.041) & (0.045) \\ [1ex]
        \textbf{Random Effects} ($\mathbf{\sigma^2}$) & & \\ [1ex]
        Pair ID Intercept & 0.076 & 0.062 & 0.073 \\ 
        Residual & 0.899 & 0.917 & 0.924 \\  \midrule
        Observations & 994,011 & 994,011 & 994,011 \\ 
        Log likelihood & -1,358,535 & -1,368,671 & -1,372,036 \\  \bottomrule
    \end{tabular}

    \begin{tabular}{l c c c}
        \toprule
        & \multicolumn{3}{c}{\textbf{Ovis1.6}} \\
        \cmidrule{2-4}
        & all-mpnet-base-v2 & all-distilroberta-v1 & all-MiniLM-L12-v2 \\
        \midrule
        \textbf{Fixed Effects} & & \\ [1ex]
        Intercept & -1.62$^{***}$ & -2.12$^{***}$ & -2.07$^{***}$ \\ 
         & (0.38) & (0.32) & (0.36) \\ [1ex]
        Gender (Women) & 0.50$^{***}$ & 0.60$^{***}$ & 0.37$^{***}$ \\ 
         & (0.053) & (0.045) & (0.050) \\ [1ex]
        Prototypicality & 0.22$^{***}$ & 0.29$^{***}$ & 0.30$^{***}$ \\ 
         & (0.058) & (0.049) & (0.054) \\ [1ex]
        \textbf{Random Effects} ($\mathbf{\sigma^2}$) & & \\ [1ex]
        Pair ID Intercept & 0.12 & 0.09 & 0.11 \\ 
        Residual & 0.84 & 0.87 & 0.88 \\  \midrule
        Observations & 999,000 & 999,000 & 999,000 \\ 
        Log likelihood & -1,334,519 & -1,347,937 & -1,355,616 \\  \bottomrule
    \end{tabular}
\end{table*}

\begin{table*}[!htbp]
    \caption{Summary output of Gender and Prototypicality models (\textbf{BLIP-3} and \textbf{Llama-3.2}). A significantly positive Gender term indicates that cosine similarity of women was notably greater than men, and a significantly positive Prototypicality term indicates a positive relationship between mean gender prototypicality and cosine similarity.}
    \label{Table: Gender and Protototypicality Models (BLIP-3, Llama-3.2)}
    \centering
    \vspace{1em}
    
    \begin{tabular}{l c c c}
        \toprule
        & \multicolumn{3}{c}{\textbf{BLIP-3}} \\
        \cmidrule{2-4}
        & all-mpnet-base-v2 & all-distilroberta-v1 & all-MiniLM-L12-v2 \\
        \midrule
        \textbf{Fixed Effects} & & \\ [1ex]
        Intercept & -1.43$^{***}$ & -1.12$^{***}$ & -1.21$^{***}$ \\ 
         & (0.22) & (0.23) & (0.27) \\ [1ex]
        Gender (Women) & 0.49$^{***}$ & 0.19$^{***}$ & 0.23$^{***}$ \\ 
         & (0.031) & (0.032) & (0.037) \\ [1ex]
        Prototypicality & 0.19$^{***}$ & 0.17$^{***}$ & 0.18$^{***}$ \\ 
         & (0.033) & (0.035) & (0.040) \\ [1ex]
        \textbf{Random Effects} ($\mathbf{\sigma^2}$) & & \\ [1ex]
        Pair ID Intercept & 0.040 & 0.044 & 0.059 \\ 
        Residual & 0.928 & 0.954 & 0.939 \\  \midrule
        Observations & 955,510 & 955,510 & 955,510 \\ 
        Log likelihood & -1,320,954 & -1,334,469 & -1,326,543 \\  \bottomrule
    \end{tabular}

    \begin{tabular}{l c c c}
        \toprule
        & \multicolumn{3}{c}{\textbf{Llama-3.2}} \\
        \cmidrule{2-4}
        & all-mpnet-base-v2 & all-distilroberta-v1 & all-MiniLM-L12-v2 \\
        \midrule
        \textbf{Fixed Effects} & & \\ [1ex]
        Intercept & -0.82 & -1.99$^{***}$ & -1.46$^{**}$ \\ 
         & (0.46) & (0.50) & (0.49) \\ [1ex]
        Gender (Women) & 0.72$^{***}$ & 0.82$^{***}$ & 0.70$^{***}$ \\ 
         & (0.064) & (0.070) & (0.068) \\ [1ex]
        Prototypicality & 0.075 & 0.26$^{***}$ & 0.18$^{*}$ \\ 
         & (0.070) & (0.076) & (0.074) \\ [1ex]
        \textbf{Random Effects} ($\mathbf{\sigma^2}$) & & \\ [1ex]
        Pair ID Intercept & 0.18 & 0.21 & 0.20 \\ 
        Residual & 0.72 & 0.68 & 0.72 \\  \midrule
        Observations & 933,168 & 933,168 & 933,168 \\ 
        Log likelihood & -1,169,502 & -1,147,999 & -1,174,777 \\  \bottomrule
    \end{tabular}
    
\end{table*}


\begin{table*}[!htbp]
    \caption{Summary output of the Gender Interaction models (\textbf{GPT-4 Turbo} and \textbf{Ovis1.6}). A significantly positive Interaction term indicates that the relationship between mean gender prototypicality and cosine similarity of women is significantly greater than that of men.}
    \label{Table: Gender Interaction Models (GPT-4 Turbo, Ovis1.6)}
    \centering
    \footnotesize
    \vspace{1em}
    
    \begin{tabular}{l c c c}
        \toprule
        & \multicolumn{3}{c}{\textbf{GPT-4 Turbo}} \\
        \cmidrule{2-4}
        & all-mpnet-base-v2 & all-distilroberta-v1 & all-MiniLM-L12-v2 \\
        \midrule
        \textbf{Fixed Effects} & & \\ [1ex]
        Intercept & -0.30 & -0.26 & 0.34 \\ 
         & (0.82) & (0.74) & (0.80) \\ [1ex]
        Gender (Women) & -2.33$^{**}$ & -2.72$^{***}$ & -2.34$^{**}$ \\ 
         & (0.87) & (0.78) & (0.85) \\ [1ex]
        Prototypicality & 0.03 & 0.03 & -0.050 \\ 
         & (0.12) & (0.11) & (0.12) \\ [1ex]
        Interaction & 0.44$^{**}$ & 0.49$^{***}$ & 0.39$^{**}$ \\ 
         & (0.13) & (0.12) & (0.13) \\ [1ex]
        \textbf{Random Effects} ($\mathbf{\sigma^2}$) & & \\ [1ex]
        Pair ID Intercept & 0.074 & 0.060 & 0.071 \\ 
        Residual & 0.899 & 0.917 & 0.924 \\  \midrule
        Observations & 994,011 & 994,011 & 994,011 \\ 
        Log likelihood & -1,358,531 & -1,368,664 & -1,372,033 \\  \bottomrule
    \end{tabular}

    \begin{tabular}{l c c c}
        \toprule
        & \multicolumn{3}{c}{\textbf{Ovis1.6}} \\
        \cmidrule{2-4}
        & all-mpnet-base-v2 & all-distilroberta-v1 & all-MiniLM-L12-v2 \\
        \midrule
        \textbf{Fixed Effects} & & \\ [1ex]
        Intercept & -4.46$^{***}$ & -3.80$^{***}$ & -4.38$^{***}$ \\ 
         & (1.04) & (0.88) & (0.97) \\ [1ex]
        Gender (Women) & 3.74$^{***}$ & 2.51$^{**}$ & 2.99$^{**}$ \\ 
         & (1.10) & (0.94) & (1.03) \\ [1ex]
        Prototypicality & 0.65$^{***}$ & 0.55$^{***}$ & 0.66$^{***}$ \\ 
         & (0.16) & (0.13) & (0.15) \\ [1ex]
        Interaction & -0.50$^{**}$ & -0.29$^{*}$ & -0.40$^{*}$ \\ 
         & (0.17) & (0.14) & (0.16) \\ [1ex]
        \textbf{Random Effects} ($\mathbf{\sigma^2}$) & & \\ [1ex]
        Pair ID Intercept & 0.12 & 0.09 & 0.10 \\ 
        Residual & 0.84 & 0.87 & 0.88 \\  \midrule
        Observations & 999,000 & 999,000 & 999,000 \\ 
        Log likelihood & -1,334,515 & -1,347,936 & -1,355,614 \\  \bottomrule
    \end{tabular}
    
\end{table*}

\begin{table*}[!hbtp]
    \caption{Summary output of the Gender Interaction models (\textbf{BLIP-3} and \textbf{Llama-3.2}). A significantly positive Interaction term indicates that the relationship between mean gender prototypicality and cosine similarity of women is significantly greater than that of men.}
    \label{Table: Gender Interaction Models (BLIP-3, Llama-3.2)}
    \footnotesize
    \vspace{1em}
    \centering

    \begin{tabular}{l c c c}
        \toprule
        & \multicolumn{3}{c}{\textbf{BLIP-3}} \\
        \cmidrule{2-4}
        & all-mpnet-base-v2 & all-distilroberta-v1 & all-MiniLM-L12-v2 \\
        \midrule
        \textbf{Fixed Effects} & & \\ [1ex]
        Intercept & -0.33 & -0.15 & -0.10 \\ 
         & (0.60) & (0.63) & (0.73) \\ [1ex]
        Gender & -0.76 & -0.91 & -1.03 \\ 
         & (0.64) & (0.67) & (0.77) \\ [1ex]
        Prototypicality & 0.026 & 0.020 & 0.0089 \\ 
         & (0.092) & (0.096) & (0.111) \\ [1ex]
        Interaction & 0.19$^{.}$ & 0.17 & 0.19 \\ 
         & (0.098) & (0.103) & (0.119) \\ [1ex]
        \textbf{Random Effects} ($\mathbf{\sigma^2}$) & & \\ [1ex]
        Pair ID Intercept & 0.040 & 0.044 & 0.059 \\ 
        Residual & 0.928 & 0.954 & 0.939 \\  \midrule
        Observations & 955,510 & 955,510 & 955,510 \\ 
        Log likelihood & -1,320,954 & -1,334,469 & -1,326,542 \\  \bottomrule
    \end{tabular}
    
    \begin{tabular}{l c c c}
        \toprule
        & \multicolumn{3}{c}{\textbf{Llama-3.2}} \\
        \cmidrule{2-4}
        & all-mpnet-base-v2 & all-distilroberta-v1 & all-MiniLM-L12-v2 \\
        \midrule
        \textbf{Fixed Effects} & & \\ [1ex]
        Intercept & 3.62$^{**}$ & 2.41$^{.}$ & 3.00$^{*}$ \\ 
         & (1.24) & (1.37) & (1.32) \\ [1ex]
        Gender & -4.33$^{**}$ & -4.18$^{**}$ & -4.37$^{**}$ \\ 
         & (1.31) & (1.45) & (1.40) \\ [1ex]
        Prototypicality & -0.60$^{**}$ & -0.41$^{*}$ & -0.50$^{*}$ \\ 
         & (0.19) & (0.21) & (0.20) \\ [1ex]
        Interaction & 0.78$^{***}$ & 0.77$^{***}$ & 0.78$^{***}$ \\ 
         & (0.20) & (0.22) & (0.22) \\ [1ex]
        \textbf{Random Effects} ($\mathbf{\sigma^2}$) & & \\ [1ex]
        Pair ID Intercept & 0.17 & 0.21 & 0.19 \\ 
        Residual & 0.72 & 0.68 & 0.72 \\  \midrule
        Observations & 933,168 & 933,168 & 933,168 \\ 
        Log likelihood & -1,169,495 & -1,147,994 & -1,174,771 \\  \bottomrule
    \end{tabular}
    
\end{table*}


\begin{table*}[!htbp]
    \caption{Likelihood-ratio test results (Gender). Significant $\chi^2$ statistic indicates that the effect of interest improved model fit.}
    \label{Table: LRT (Gender)}
    \vspace{1em}
    \footnotesize
    \centering
    
    \begin{tabular}{clllccc}
        \toprule
        \textbf{VLM} & \textbf{Encoder Model} & \textbf{Mixed-Effects Model} & \textbf{Effect} & $\mathbf{\chi^2}$ & \textbf{df} & \textbf{\textit{p}} \\ \midrule
        & & Gender and Prototypicality model & Gender & 133.82 & 1 & < .001 \\
        & all-mpnetbase-v2 & Gender and Prototypicality model & Prototypicality & 74.86 & 1 & < .001 \\
        & & Interaction model & Interaction & 138.03 & 2 & < .001 \\ \cmidrule{2-7}
        & & Gender and Prototypicality model & Gender & 128.12 & 1 & < .001 \\
        GPT-4 Turbo & all-distilroberta-v1 & Gender and Prototypicality model & Prototypicality & 107.12 & 1 & < .001 \\
        & & Interaction model & Interaction & 137.44 & 2 & < .001 \\ \cmidrule{2-7}
        & & Gender and Prototypicality model& Gender & 21.84 & 1 & < .001 \\
        & all-MiniLM-L12-v2 & Gender and Prototypicality model & Prototypicality & 40.13 & 1 & < .001 \\
        & & Interaction model & Interaction & 41.58 & 2 & < .001 \\
        \bottomrule

        \midrule
        \textbf{VLM} & \textbf{Encoder Model} & \textbf{Mixed-Effects Model} & \textbf{Effect} & $\mathbf{\chi^2}$ & \textbf{df} & \textbf{\textit{p}} \\ \midrule
        & & Gender and Prototypicality model & Gender & 81.02 & 1 & < .001 \\
        & all-mpnetbase-v2 & Gender and Prototypicality model & Prototypicality & 14.46 & 1 & < .001 \\
        & & Interaction model & Interaction & 100.94 & 2 & < .001 \\ \cmidrule{2-7}
        & & Gender and Prototypicality model & Gender & 147.48 & 1 & < .001 \\
        Ovis1.6 & all-distilroberta-v1 & Gender and Prototypicality model & Prototypicality & 34.76 & 1 & < .001 \\
        & & Interaction model & Interaction & 173.32 & 2 & < .001 \\ \cmidrule{2-7}
        & & Gender and Prototypicality model & Gender & 52.37 & 1 & < .001 \\
        & all-MiniLM-L12-v2 & Gender and Prototypicality model & Prototypicality & 30.89 & 1 & < .001 \\
        & & Interaction model & Interaction & 50.44 & 2 & < .001 \\ \bottomrule

        \midrule
        \textbf{VLM} & \textbf{Encoder Model} & \textbf{Mixed-Effects Model} & \textbf{Effect} & $\mathbf{\chi^2}$ & \textbf{df} & \textbf{\textit{p}} \\ \midrule
        & & Gender and Prototypicality model & Gender & 196.76 & 1 & < .001 \\
        & all-mpnetbase-v2 & Gender and Prototypicality model & Prototypicality & 32.00 & 1 & < .001 \\
        & & Interaction model & Interaction & 257.05 & 1 & < .001 \\ \cmidrule{2-7}
        & & Gender and Prototypicality model & Gender & 32.38 & 1 & < .001 \\
        BLIP-3 & all-distilroberta-v1 & Gender and Prototypicality model & Prototypicality & 22.07 & 1 & < .001 \\
        & & Interaction model & Interaction & 33.44 & 2 & < .001 \\ \cmidrule{2-7}
        & & Gender and Prototypicality model & Gender & 36.99 & 1 & < .001 \\
        & all-MiniLM-L12-v2 & Gender and Prototypicality model & Prototypicality & 19.06 & 1 & < .001 \\
        & & Interaction model & Interaction & 38.16 & 2 & < .001 \\ \bottomrule

        \midrule
        \textbf{VLM} & \textbf{Encoder Model} & \textbf{Mixed-Effects Model} & \textbf{Effect} & $\mathbf{\chi^2}$ & \textbf{df} & \textbf{\textit{p}} \\ \midrule
        & & Gender and Prototypicality model & Gender & 112.31 & 1 & < .001 \\
        & all-mpnetbase-v2 & Gender and Prototypicality model & Prototypicality & 1.18 & 1 & .28 \\
        & & Interaction model & Interaction & 212.22 & 2 & < .001 \\ \cmidrule{2-7}
        & & Gender and Prototypicality model & Gender & 119.44 & 1 & < .001 \\
        Llama-3.2 & all-distilroberta-v1 & Gender and Prototypicality model & Prototypicality & 11.14 & 1 & < .001 \\
        & & Interaction model & Interaction & 177.82 & 2 & < .001 \\ \cmidrule{2-7}
        & & Gender and Prototypicality model & Gender & 94.09 & 1 & < .001 \\
        & all-MiniLM-L12-v2 & Gender and Prototypicality model & Prototypicality & 6.01 & 1 & < .001 \\
        & & Interaction model & Interaction & 150.88 & 2 & < .001 \\ \bottomrule
        
    \end{tabular}
\end{table*}

\clearpage
\newpage

\begin{figure*}[!htbp]
    \centering
    \begin{subfigure}{\textwidth}
        \centering
        \includegraphics[width=\textwidth]{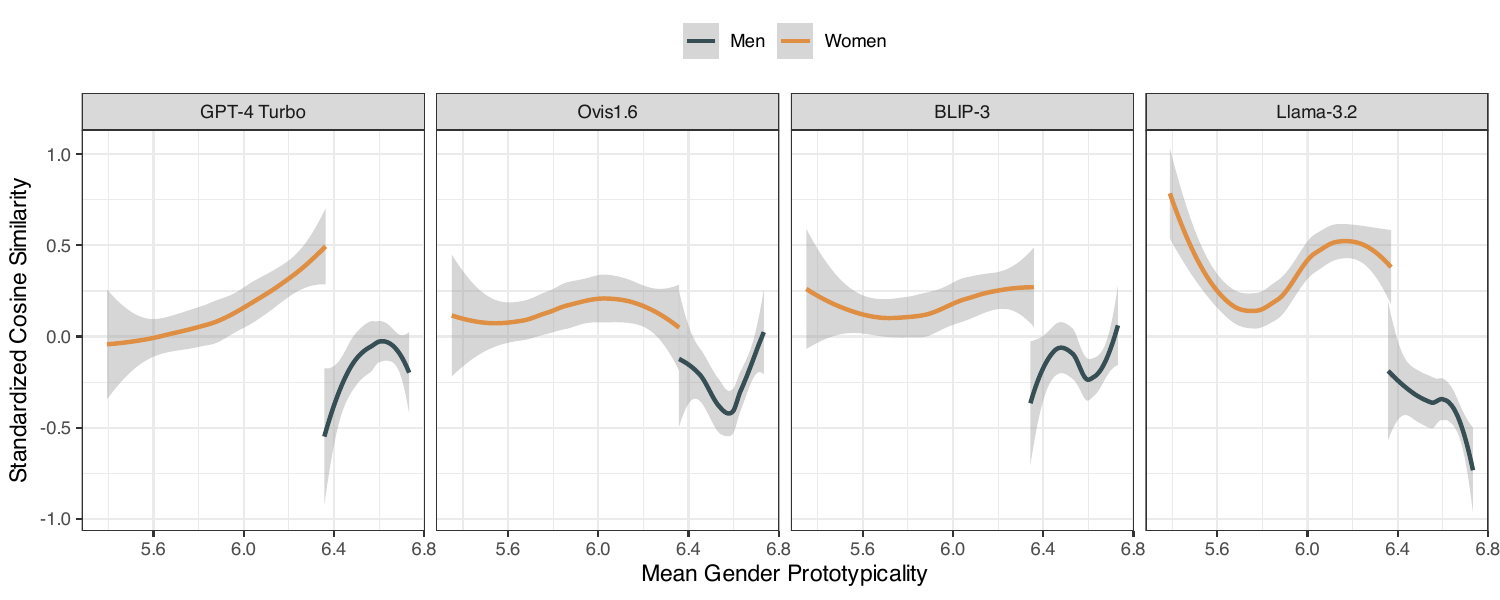}
    \end{subfigure}
    \hfill
    \begin{subfigure}{\textwidth}
        \centering
        \includegraphics[width=\textwidth]{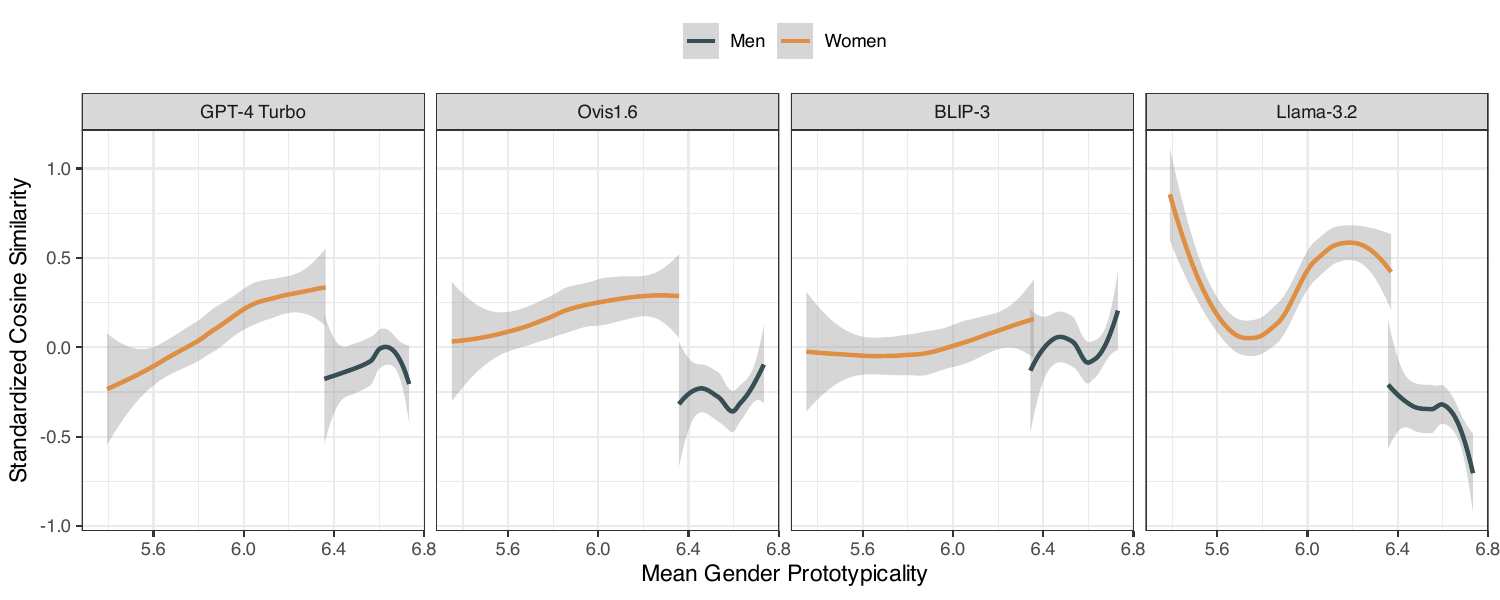}
    \end{subfigure}
    \hfill
    \begin{subfigure}{\textwidth}
        \centering
        \includegraphics[width=\textwidth]{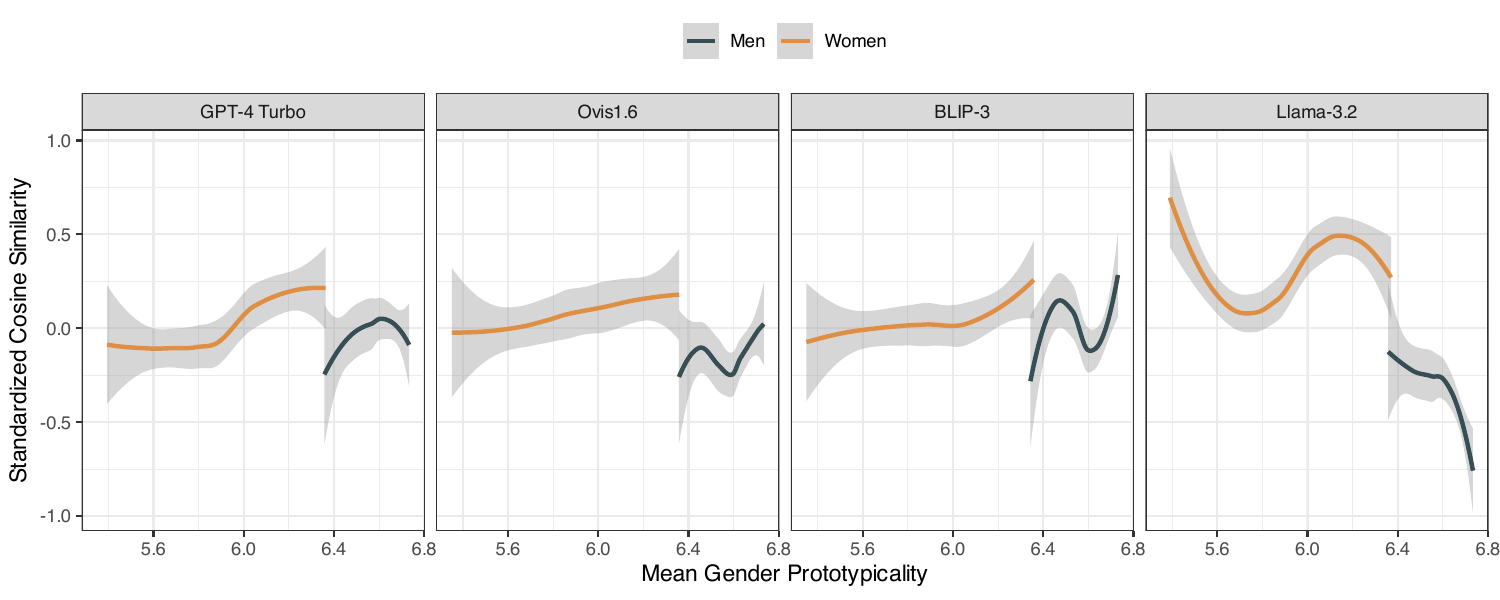}
    \end{subfigure}
    \caption{Standardized cosine similarity (1,000 random samples for each gender group) by prototypicality, calculated using all three encoder models. The top and bottom 10\% of prototypicality values were excluded to minimize tail effects.}
    \label{Figure: Gender Interactions (Supplement)}
\end{figure*}


\begin{table*}[!htbp]
    \caption{STM output table containing the expected proportion of each topic (\%), the top three FREX words, words that are both frequent and exclusive to each topic identified by the STM, and example texts by topic.}
    \label{Table: STM Topics (Gender)}
    \centering
    \footnotesize
    \vspace{1em}
    
    \begin{tabular}{C{0.06\textwidth} C{0.12\textwidth} C{0.12\textwidth} C{0.12\textwidth} m{0.44\textwidth}}
    \toprule
    \textbf{Topic} & \textbf{Proportions} & \textbf{FREX Words} & \textbf{Topic Label} & \textbf{Example} \\ \midrule
    1 & 9.29\% & local, communiti, share, school, basketbal & Basketball & On a typical day, our protagonist rises early, driven by a passion for basketball. After a quick breakfast, he heads to school, where he excels academically and plays on the varsity team. In the afternoons, he practices at the local court, honing his skills. As the sun sets, he enjoys time with friends, discussing future dreams and aspirations. The day concludes with quiet reflection, fueling his commitment to excellence and dreams of a brighter future. \\ \midrule
    2 & 14.40\% & help, gentl, reflect, began, vibrant & -- & In the crisp morning light, Sarah awoke in her cozy Brooklyn apartment, a blend of soft city sounds greeting her day. She dressed in her favorite light gray scarf, preparing for a day filled with the creative challenges of her graphic design job. Her commute was a gentle reminder of the city's vibrant energy. At the studio, she immersed herself in crafting digital art, surrounded by the hum of creativity and collaboration. As the day wound down, she cherished the quiet moments, reflecting on the joy of turning ideas into reality. \\ \midrule
    3 & 15.36\% & dinner, project, good, gym, break & -- & As a freelance writer, Emma starts her day with a cup of coffee and a quick review of her schedule. She spends the morning working on articles, responding to client emails, and brainstorming new ideas. In the afternoon, she heads to the gym for a workout, followed by a relaxing yoga session. Dinner and a movie with her friends mark the end of her day, before she finally calls it a night. \\ \midrule
    4 & 19.89\% & jazz, novel, paint, sketch, everi & Art & Amidst bustling New York streets, he brews morning coffee, savoring its aromatic warmth. He jogs through Central Park, embracing the rhythm of city life. At work, creativity flourishes within digital confines. Evenings find him strumming guitar chords, blending melodies with dreams before nightfall embraces another bustling tomorrow. \\ \midrule
    5 & 8.33\% & alarm, stretch, rub, yawn, kitchen & -- & The young man woke up to the sound of his alarm blaring in his ear. He rubbed the sleep from his eyes and swung his legs over the side of the bed. After a quick shower, he grabbed a granola bar and headed out the door to catch the bus to work. \\ \midrule
    6 & 8.95\% & patient, lab, care, hospital, nurs & Healthcare & As a nurse, Jenna starts her day early, waking up at 5am to prepare for a 12-hour shift at the hospital. She puts on her scrubs and begins her rounds, checking on patients and administering medication. Throughout the day, she remains focused and compassionate, providing comfort and care to those in need. After her shift, Jenna returns home to her husband and two children, grateful for the opportunity to make a difference in people's lives. \\ \midrule
    7 & 17.42\% & thought, lost, gaze, rose, sip & -- & As she sipped her coffee, Sarah gazed out the window, lost in thought. The sun rose over the city, casting a warm glow over the bustling streets. She took a deep breath, feeling the stress melt away. Today was a new day, full of possibilities and promise. She smiled, ready to take on whatever came her way. \\ \midrule
    8 & 6.37\% & hair, shirt, cur, beard, woman & Appearance & A young woman with short dark hair and a shaved headline. She's wearing a white shirt and has a serious expression on her face. She's standing in a white background, staring off into the distance. She's not smiling, but her expression is calm and composed. \\
     \bottomrule
    \end{tabular}
\end{table*}

\begin{table*}[!htbp]
    \caption{The effect of gender, gender prototypicality, and their interactions in individual VLMs across the five topics. A significant positive gender term indicates that the topic is significantly more prevalent for women than men, a significant positive prototypicality term indicates that a unit increase in prototypicality is associated with a significant increase in topic prevalence, and a significant interaction term indicates that the effect of prototypicality on topic prevalence is significantly greater for women than men.}
    \label{Table: Trait Associations (Gender)}
    \centering
    \footnotesize
    \vspace{1em}
    
    \begin{tabular}{l cccc}
    
        \toprule
        & \multicolumn{4}{c}{\textbf{Gender}} \\ \cmidrule{2-5}
        & \textbf{Basketball} & \textbf{Art} & \textbf{Healthcare} & \textbf{Appearance} \\ \midrule 
        GPT-4 Turbo & 0.014$^{*}$ & -0.029$^{*}$ & 0.0041 & 0.00054 \\
        & [0.00050, 0.027] & [-0.046, -0.012] & [-0.0073, 0.016] & [-0.0094, 0.010] \\ [1ex]
        Ovis1.6 & 0.0060 & -0.0037 & 0.0056 & 0.021$^{*}$ \\
        & [-0.021, 0.033] & [-0.029, 0.022] & [-0.015, 0.026] & [0.0034, 0.038] \\ [1ex]
        BLIP-3 & -0.0044 & 0.00032 & 0.041$^{*}$ & 0.000052 \\
        & [-0.027, 0.018] & [-0.025, 0.026] & [0.017, 0.066] & [-0.018, 0.019] \\ [1ex]
        Llama-3.2 & -0.0094 & 0.0094 & 0.014 & -0.022$^{*}$ \\
        & [-0.033, 0.014] & [-0.016, 0.035] & [-0.0058, 0.034] & [-0.039, -0.0042] \\ \bottomrule 
    
        \midrule
        & \multicolumn{4}{c}{\textbf{Interaction}} \\ \cmidrule{2-5}
        & \textbf{Basketball} & \textbf{Art} & \textbf{Healthcare} & \textbf{Appearance} \\ \midrule 
        GPT-4 Turbo & -0.0088 & 0.0042 & 0.0020 & 0.013 \\
        & [-0.053, 0.036] & [-0.043, 0.051] & [-0.039, 0.043] & [-0.023, 0.049] \\ [1ex]
        Ovis1.6 & 0.096 & -0.012 & 0.026 & 0.026 \\
        & [0.015, 0.18] & [-0.086, 0.062] & [-0.048, 0.10] & [-0.037, 0.089] \\ [1ex]
        BLIP-3 & 0.029 & 0.0015 & -0.027 & 0.043 \\
        & [-0.048, 0.11] & [-0.071, 0.074] & [-0.12, 0.064] & [-0.033, 0.12] \\ [1ex]
        Llama-3.2 & -0.0059 & 0.0067 & 0.0078 & -0.0048 \\
        & [-0.081, 0.069] & [-0.066, 0.080] & [-0.064, 0.080] & [-0.068, 0.059] \\ \bottomrule 
    
    \end{tabular}
\end{table*}

\clearpage
\newpage

\begin{figure*}[!htbp]
    \centering
    \includegraphics[width=\textwidth]{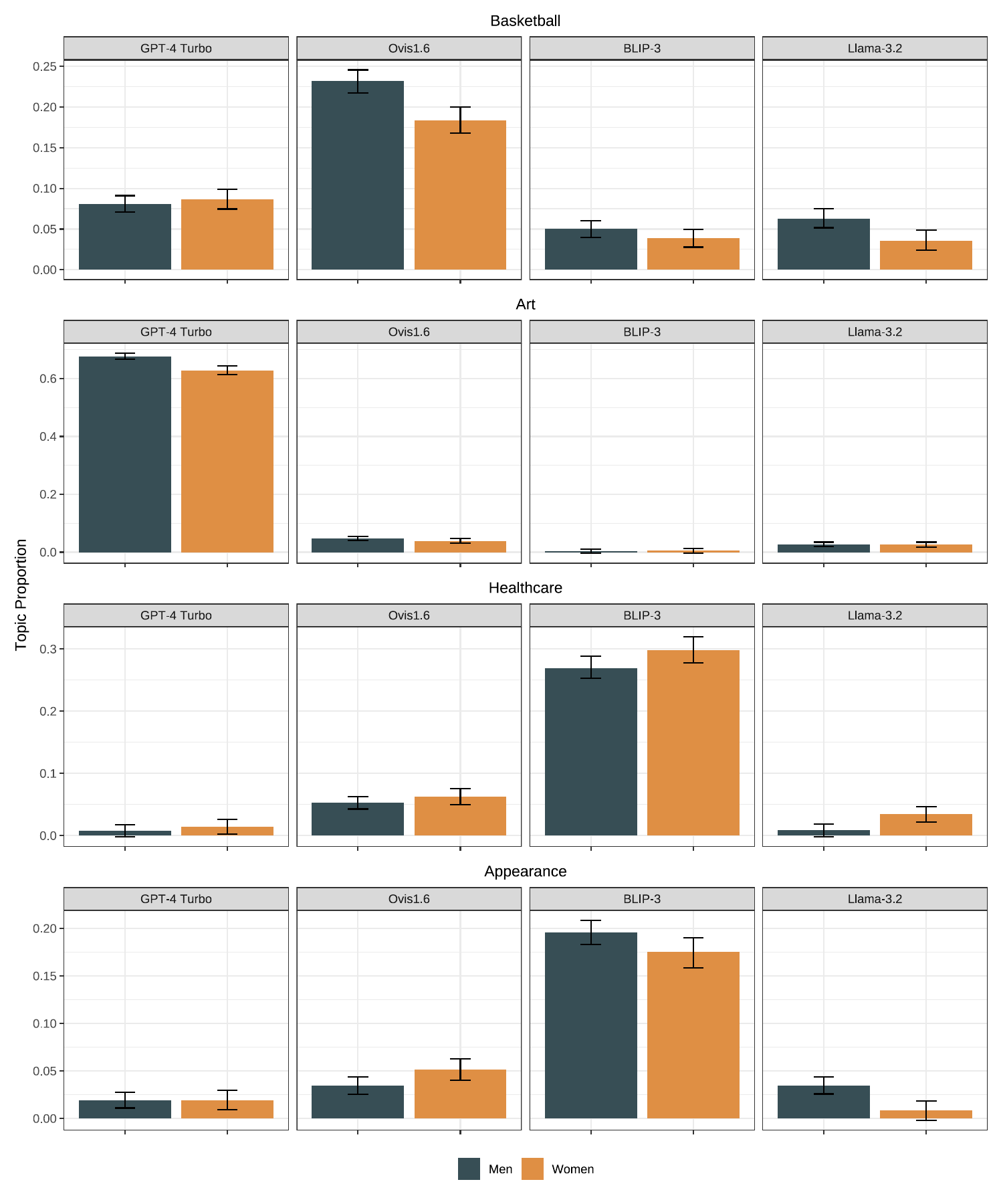}
    \caption{Prevalence of all four topics across four VLMs. Error bars indicate 95\% confidence intervals.}
    \label{Figure: Trait Associations (Gender)}
\end{figure*}

\clearpage
\newpage

\section{Results (Race)}
\label{Appendix: Race Results}

\begin{table*}[!htbp]
    \caption{Summary output of the Race and Prototypicality models (\textbf{GPT-4 Turbo} and \textbf{Ovis1.6}). A significantly positive Race term indicates that cosine similarity of Black Americans was notably greater than White Americans. A significantly positive Prototypicality term indicates a positive relationship between mean prototypicality and cosine similarity.}
    \label{Table: Race and Prototypicality Models (GPT-4 Turbo, Ovis1.6)}
    \vspace{1em}
    \centering
    \begin{tabular}{l c c c}
        \toprule
        & \multicolumn{3}{c}{\textbf{GPT-4 Turbo}} \\ \cmidrule{2-4}
        & all-mpnet-base-v2 & all-distilroberta-v1 & all-MiniLM-L12-v2 \\
        \midrule
        \textbf{Fixed Effects} & & \\ [1ex]
        Intercept & -0.26 & 0.040 & 0.18 \\ 
         & (0.22) & (0.25) & (0.19) \\ [1ex]
        Race & -0.46$^{***}$ & -0.36$^{***}$ & -0.44$^{***}$ \\ 
         & (0.038) & (0.043) & (0.032) \\ [1ex]
        Prototypicality & 0.084$^{*}$ & 0.027 & 0.011 \\ 
         & (0.036) & (0.041) & (0.030) \\ [1ex]
        \textbf{Random Effects} ($\mathbf{\sigma^2}$) & & \\ [1ex]
        Pair ID Intercept & 0.15 & 0.19 & 0.10 \\ 
        Residual & 0.80 & 0.78 & 0.85 \\  \midrule
        Observations & 995,010 & 995,010 & 995,010 \\ 
        Log likelihood & -1302151.00 & -1292041.00 & -1334234.00 \\  \bottomrule
    \end{tabular}

    \begin{tabular}{l c c c}
        \toprule
        & \multicolumn{3}{c}{\textbf{Ovis1.6}} \\
        \cmidrule{2-4}
        & all-mpnet-base-v2 & all-distilroberta-v1 & all-MiniLM-L12-v2 \\
        \midrule
        \textbf{Fixed Effects} & & \\ [1ex]
        Intercept & 0.54$^{.}$ & 0.086 & 0.76$^{**}$ \\ 
         & (0.28) & (0.35) & (0.29) \\ [1ex]
        Race & 0.0021 & 0.030 & -0.081 \\ 
         & (0.047) & (0.059) & (0.048) \\ [1ex]
        Prototypicality & -0.085$^{.}$ & -0.011 & -0.11$^{*}$ \\ 
         & (0.045) & (0.056) & (0.046) \\ [1ex]
        \textbf{Random Effects} ($\mathbf{\sigma^2}$) & & \\ [1ex]
        Pair ID Intercept & 0.23 & 0.36 & 0.24 \\ 
        Residual & 0.78 & 0.65 & 0.76 \\  \midrule
        Observations & 999,000 & 999,000 & 999,000 \\ 
        Log likelihood & -1,292,578 & -1,202,381 & -1,282,517 \\  \bottomrule
    \end{tabular}
    
\end{table*}

\begin{table*}[!htbp]
    \caption{Summary output of the Race and Prototypicality models (\textbf{BLIP-3} and \textbf{Llama-3.2}). A significantly positive Race term indicates that cosine similarity of Black Americans was notably greater than White Americans. A significantly positive Prototypicality term indicates a positive relationship between mean prototypicality and cosine similarity.}
    \label{Table: Race and Prototypicality Models (BLIP-3, Llama-3.2)}
    \vspace{1em}
    \centering
    \begin{tabular}{l c c c}
        \toprule
        & \multicolumn{3}{c}{\textbf{BLIP-3}} \\
        \cmidrule{2-4}
        & all-mpnet-base-v2 & all-distilroberta-v1 & all-MiniLM-L12-v2 \\
        \midrule
        \textbf{Fixed Effects} & & \\ [1ex]
        Intercept & 0.38 & 0.43 & 0.51$^{*}$ \\ 
         & (0.21) & (0.24) & (0.23) \\ [1ex]
        Race & -0.17$^{***}$ & -0.11$^{**}$ & -0.35$^{***}$ \\ 
         & (0.036) & (0.041) & (0.040) \\ [1ex]
        Prototypicality & -0.045 & -0.057 & -0.050 \\ 
         & (0.034) & (0.039) & (0.038) \\ [1ex]
        \textbf{Random Effects} ($\mathbf{\sigma^2}$) & & \\ [1ex]
        Pair ID Intercept & 0.13 & 0.18 & 0.16 \\ 
        Residual & 0.86 & 0.82 & 0.81 \\  \midrule
        Observations & 955,627 & 955,627 & 955,627 \\ 
        Log likelihood & -1,287,785 & -1,265,156 & -1,257,435 \\  \bottomrule
    \end{tabular}

    \begin{tabular}{l c c c}
        \toprule
        & \multicolumn{3}{c}{\textbf{Llama-3.2}} \\ \cmidrule{2-4}
        & all-mpnet-base-v2 & all-distilroberta-v1 & all-MiniLM-L12-v2 \\
        \midrule
        \textbf{Fixed Effects} & & \\ [1ex]
        Intercept & 0.70$^{*}$ & 1.08$^{***}$ & 0.78$^{*}$ \\ 
         & (0.31) & (0.31) & (0.36) \\ [1ex]
        Race & -0.15$^{**}$ & -0.27$^{***}$ & -0.097 \\ 
         & (0.052) & (0.053) & (0.061) \\ [1ex]
        Prototypicality & -0.10$^{*}$ & -0.15$^{**}$ & -0.11$^{*}$ \\ 
         & (0.049) & (0.051) & (0.058) \\ [1ex]
        \textbf{Random Effects} ($\mathbf{\sigma^2}$) & & \\ [1ex]
        Pair ID Intercept & 0.28 & 0.30 & 0.39 \\ 
        Residual & 0.72 & 0.69 & 0.62 \\  \midrule
        Observations & 933,396 & 933,396 & 933,396 \\ 
        Log likelihood & -1,172,468 & -1,155,491 & -1,100,526 \\  \bottomrule
    \end{tabular}
    
\end{table*}


\begin{table*}[!htbp]
    \caption{Summary output of the Race Interaction models (\textbf{GPT-4 Turbo} and \textbf{Ovis1.6}). A significantly positive Interaction term indicates that the relationship between mean prototypicality and cosine similarity of Black Americans is significantly greater than that of White Americans.}
    \label{Table: Race Interaction Models (GPT-4 Turbo, Ovis1.6)}
    \vspace{1em}
    \centering
    
    \begin{tabular}{l c c c}
        \toprule
        & \multicolumn{3}{c}{\textbf{GPT-4 Turbo}} \\
        \cmidrule{2-4}
        & all-mpnet-base-v2 & all-distilroberta-v1 & all-MiniLM-L12-v2 \\
        \midrule
        \textbf{Fixed Effects} & & \\ [1ex]
        Intercept & -0.20 & -0.40 & -0.98 \\ 
         & (0.56) & (0.63) & (0.46) \\ [1ex]
        Race (Black) & -0.53 & 0.17 & 0.93 \\ 
         & (0.61) & (0.69) & (0.50) \\ [1ex]
        Prototypicality & 0.074 & 0.098 & 0.20$^{**}$ \\ 
         & (0.090) & (0.10) & (0.074) \\ [1ex]
        Interaction & 0.011 & -0.085 & -0.22$^{**}$ \\ 
         & (0.098) & (0.11) & (0.081) \\ [1ex]
        \textbf{Random Effects} ($\mathbf{\sigma^2}$) & & \\ [1ex]
        Pair ID Intercept & 0.15 & 0.19 & 0.10 \\ 
        Residual & 0.80 & 0.78 & 0.85 \\  \midrule
        Observations & 995,010 & 995,010 & 995,010 \\ 
        Log likelihood & -1302152.00 & -1292042.00 & -1334232.00 \\  \bottomrule
    \end{tabular}

    \begin{tabular}{l c c c}
        \toprule
        & \multicolumn{3}{c}{\textbf{Ovis1.6}} \\
        \cmidrule{2-4}
        & all-mpnet-base-v2 & all-distilroberta-v1 & all-MiniLM-L12-v2 \\
        \midrule
        \textbf{Fixed Effects} & & \\ [1ex]
        Intercept & 2.73$^{***}$ & 1.39 & 1.56$^{*}$ \\ 
         & (0.68) & (0.86) & (0.71) \\ [1ex]
        Race (Black) & -2.59$^{***}$ & -1.52 & -1.04 \\ 
         & (0.74) & (0.94) & (0.77) \\ [1ex]
        Prototypicality & -0.44$^{***}$ & -0.22 & -0.24$^{*}$ \\ 
         & (0.11) & (0.14) & (0.11) \\ [1ex]
        Interaction & 0.42$^{***}$ & 0.25$^{.}$ & 0.15 \\ 
         & (0.12) & (0.15) & (0.13) \\ [1ex]
        \textbf{Random Effects} ($\mathbf{\sigma^2}$) & & \\ [1ex]
        Pair ID Intercept & 0.22 & 0.36 & 0.24 \\ 
        Residual & 0.78 & 0.65 & 0.76 \\  \midrule
        Observations & 999,000 & 999,000 & 999,000 \\ 
        Log likelihood & -1,292,573 & -1,202,381 & -1,282,517 \\  \bottomrule
    \end{tabular}
\end{table*}

\begin{table*}[!hbtp]
    \caption{Summary output of the Race Interaction models (\textbf{BLIP-3} and \textbf{Llama-3.2}). A significantly positive Interaction term indicates that the relationship between mean prototypicality and cosine similarity of Black Americans is significantly greater than that of White Americans.}
    \label{Table: Race Interaction Models (BLIP-3, Llama-3.2)}
    \vspace{1em}
    \centering
    
    \begin{tabular}{l c c c}
        \toprule
        & \multicolumn{3}{c}{\textbf{BLIP-3}} \\
        \cmidrule{2-4}
        & all-mpnet-base-v2 & all-distilroberta-v1 & all-MiniLM-L12-v2 \\
        \midrule
        \textbf{Fixed Effects} & & \\ [1ex]
        Intercept & 0.38 & 0.43 & 0.51$^{*}$ \\ 
         & (0.21) & (0.24) & (0.23) \\ [1ex]
        Race (Black) & -0.17$^{***}$ & -0.11$^{**}$ & -0.35$^{***}$ \\ 
         & (0.036) & (0.041) & (0.040) \\ [1ex]
        Prototypicality & -0.045 & -0.057 & -0.050 \\ 
         & (0.034) & (0.039) & (0.038) \\ [1ex]
        Interaction & 0.19$^{*}$ & 0.13 & 0.088 \\ 
         & (0.092) & (0.11) & (0.10) \\ [1ex]
        \textbf{Random Effects} ($\mathbf{\sigma^2}$) & & \\ [1ex]
        Pair ID Intercept & 0.13 & 0.18 & 0.16 \\ 
        Residual & 0.86 & 0.82 & 0.81 \\  \midrule
        Observations & 955,627 & 955,627 & 955,627 \\ 
        Log likelihood & -1,287,785 & -1,265,157 & -1,257,436 \\  \bottomrule
    \end{tabular}
    
    \begin{tabular}{l c c c}
        \toprule
        & \multicolumn{3}{c}{\textbf{Llama-3.2}} \\
        \cmidrule{2-4}
        & all-mpnet-base-v2 & all-distilroberta-v1 & all-MiniLM-L12-v2 \\
        \midrule
        \textbf{Fixed Effects} & & \\ [1ex]
        Intercept & 0.70$^{*}$ & 0.78$^{*}$ & 1.08$^{***}$ \\ 
         & (0.31) & (0.36) & (0.31) \\ [1ex]
        Race (Black) & -0.15$^{**}$ & -0.097 & -0.27$^{***}$ \\ 
         & (0.052) & (0.061) & (0.053) \\ [1ex]
        Prototypicality & -0.10$^{*}$ & -0.11$^{*}$ & -0.15$^{**}$ \\ 
         & (0.049) & (0.058) & (0.051) \\ [1ex]
        Interaction & 0.13 & 0.13 & 0.07 \\ 
         & (0.13) & (0.16) & (0.14) \\ [1ex]
        \textbf{Random Effects} ($\mathbf{\sigma^2}$) & & \\ [1ex]
        Pair ID Intercept & 0.28 & 0.39 & 0.30 \\ 
        Residual & 0.72 & 0.62 & 0.69 \\  \midrule
        Observations & 933,396 & 933,396 & 933,396 \\ 
        Log likelihood & -1,172,469 & -1,100,526 & -1,155,492 \\  \bottomrule
    \end{tabular}
\end{table*}


\begin{table*}[!htbp]
    \caption{Likelihood-ratio test results (Race). Significant $\chi^2$ statistic indicates that the effect of interest improved model fit.}
    \label{Table: LRT (Race)}
    \vspace{1em}
    \footnotesize
    \centering
    
    \begin{tabular}{clllccc}
        \toprule
        \textbf{VLM} & \textbf{Encoder Model} & \textbf{Mixed-Effects Model} & \textbf{Effect} & $\mathbf{\chi^2}$ & \textbf{df} & \textbf{\textit{p}} \\ \midrule
        & & Race and Prototypicality model & Race & 98.16 & 1 & < .001 \\
        & all-mpnetbase-v2 & Race and Prototypicality model & Prototypicality & 1.17 & 1 & .28 \\
        & & Interaction model & Interaction & 97.67 & 2 & <.001 \\ \cmidrule{2-7}
        & & Race and Prototypicality model & Race & 13.94 & 1 & < .001 \\
        GPT-4 Turbo & all-distilroberta-v1 & Race and Prototypicality model & Prototypicality & 0.04 & 1 & .84 \\
        & & Interaction model & Interaction & 13.89 & 2 & < .001 \\ \cmidrule{2-7}
        & & Race and Prototypicality model& Race & 145.15 & 1 & < .001 \\
        & all-MiniLM-L12-v2 & Race and Prototypicality model & Prototypicality & 3.81 & 1 & .051 \\
        & & Interaction model & Interaction & 146.94 & 2 & < .001 \\ \bottomrule

        \midrule
        \textbf{VLM} & \textbf{Encoder Model} & \textbf{Mixed-Effects Model} & \textbf{Effect} & $\mathbf{\chi^2}$ & \textbf{df} & \textbf{\textit{p}} \\ \midrule
        & & Race and Prototypicality model & Race & 0.00 & 1 & .97 \\
        & all-mpnetbase-v2 & Race and Prototypicality model & Prototypicality & 3.63 & 1 & .057 \\
        & & Interaction model & Interaction & 3.57 & 2 & .15 \\ \cmidrule{2-7}
        & & Race and Prototypicality model & Race & 0.27 & 1 & .61 \\
        Ovis1.6 & all-distilroberta-v1 & Race and Prototypicality model & Prototypicality & 0.04 & 1 & .84 \\
        & & Interaction model & Interaction & 0.44 & 2 & .80 \\ \cmidrule{2-7}
        & & Race and Prototypicality model & Race & 2.84 & 1 & .092 \\
        & all-MiniLM-L12-v2 & Race and Prototypicality model & Prototypicality & 6.03 & 1 & .014 \\
        & & Interaction model & Interaction & 7.89 & 2 & 0.19 \\ \bottomrule

        \midrule
        \textbf{VLM} & \textbf{Encoder Model} & \textbf{Mixed-Effects Model} & \textbf{Effect} & $\mathbf{\chi^2}$ & \textbf{df} & \textbf{\textit{p}} \\ \midrule
        & & Race and Prototypicality model & Race & 22.17 & 1 & < .001 \\
        & all-mpnetbase-v2 & Race and Prototypicality model & Prototypicality & 1.79 & 1 & 1.81 \\
        & & Interaction model & Interaction & 21.65 & 2 & < .001 \\ \cmidrule{2-7}
        & & Race and Prototypicality model & Race & 7.66 & 1 & .006 \\
        BLIP-3 & all-distilroberta-v1 & Race and Prototypicality model & Prototypicality & 2.12 & 1 & .15 \\
        & & Interaction model & Interaction & 8.64 & 2 & .013 \\ \cmidrule{2-7}
        & & Race and Prototypicality model & Race & 70.16 & 1 & < .001 \\
        & all-MiniLM-L12-v2 & Race and Prototypicality model & Prototypicality & 1.76 & 1 & .19 \\
        & & Interaction model & Interaction & 69.17 & 2 & < .001 \\ \bottomrule

        \midrule
        \textbf{VLM} & \textbf{Encoder Model} & \textbf{Mixed-Effects Model} & \textbf{Effect} & $\mathbf{\chi^2}$ & \textbf{df} & \textbf{\textit{p}} \\ \midrule
        & & Race and Prototypicality model & Race & 7.98 & 1 & .005 \\
        & all-mpnetbase-v2 & Race and Prototypicality model & Prototypicality & 4.07 & 1 & .044 \\
        & & Interaction model & Interaction & 10.67 & 2 & .005 \\ \cmidrule{2-7}
        & & Race and Prototypicality model & Race & 2.50 & 1 & .11 \\
        Llama-3.2 & all-distilroberta-v1 & Race and Prototypicality model & Prototypicality & 3.91 & 1 & .048 \\
        & & Interaction model & Interaction & 5.70 & 2 & .058 \\ \cmidrule{2-7}
        & & Race and Prototypicality model & Race & 24.96 & 1 & < .001 \\
        & all-MiniLM-L12-v2 & Race and Prototypicality model & Prototypicality & 8.89 & 1 & .003 \\
        & & Interaction model & Interaction & 30.60 & 2 & < .001 \\ \bottomrule
        
    \end{tabular}
\end{table*}

\clearpage
\newpage

\begin{figure*}[!htbp]
    \centering
    \begin{subfigure}{\textwidth}
        \centering
        \includegraphics[width=\textwidth]{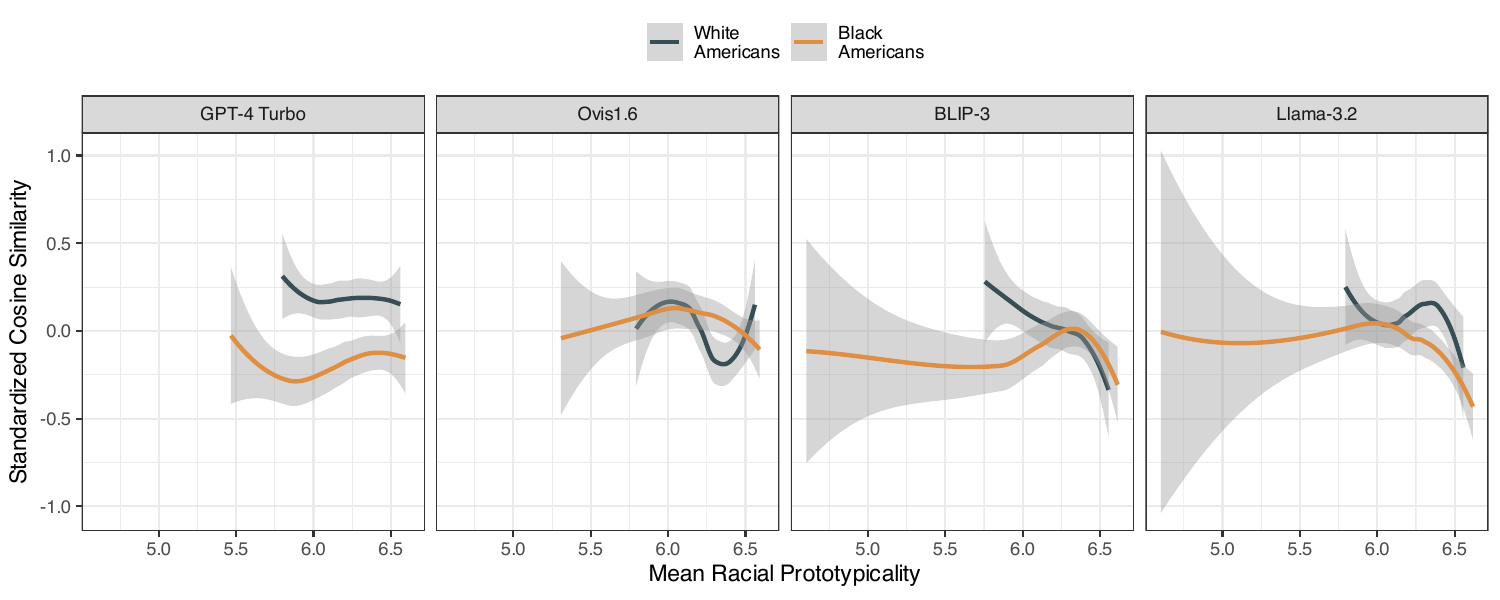}
    \end{subfigure}
    \hfill
    \begin{subfigure}{\textwidth}
        \centering
        \includegraphics[width=\textwidth]{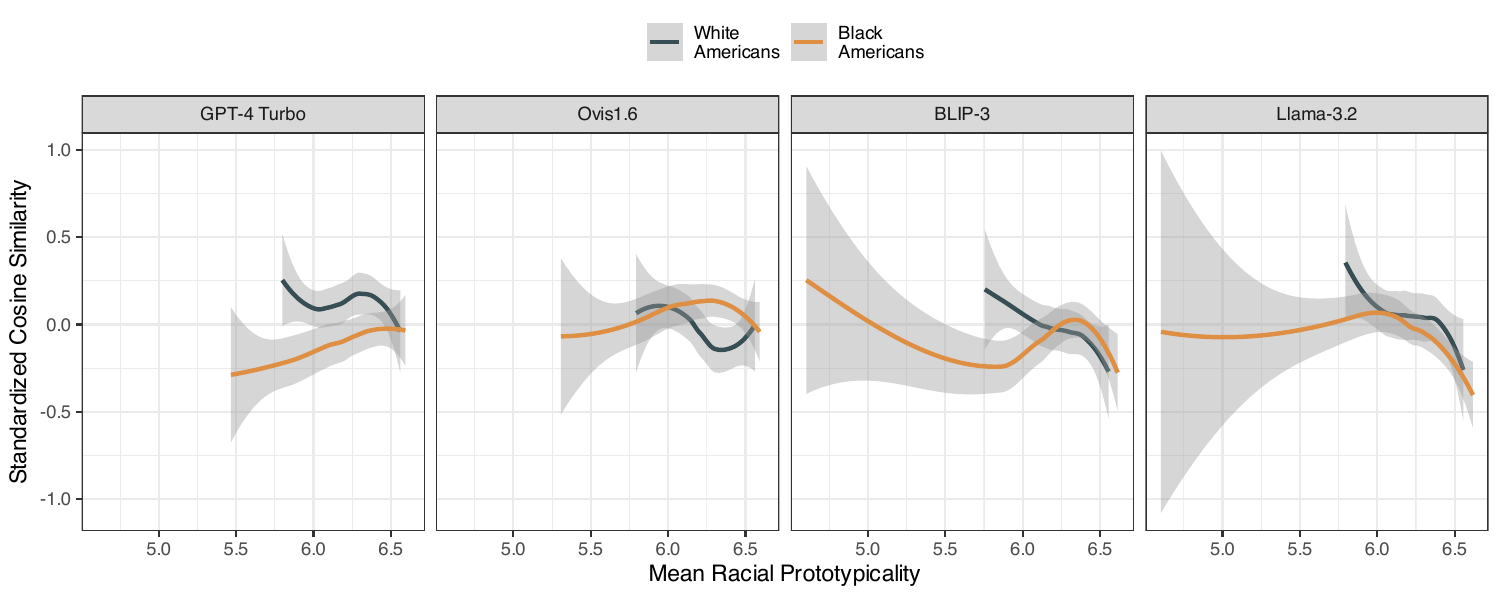}
    \end{subfigure}
    \hfill
    \begin{subfigure}{\textwidth}
        \centering
        \includegraphics[width=\textwidth]{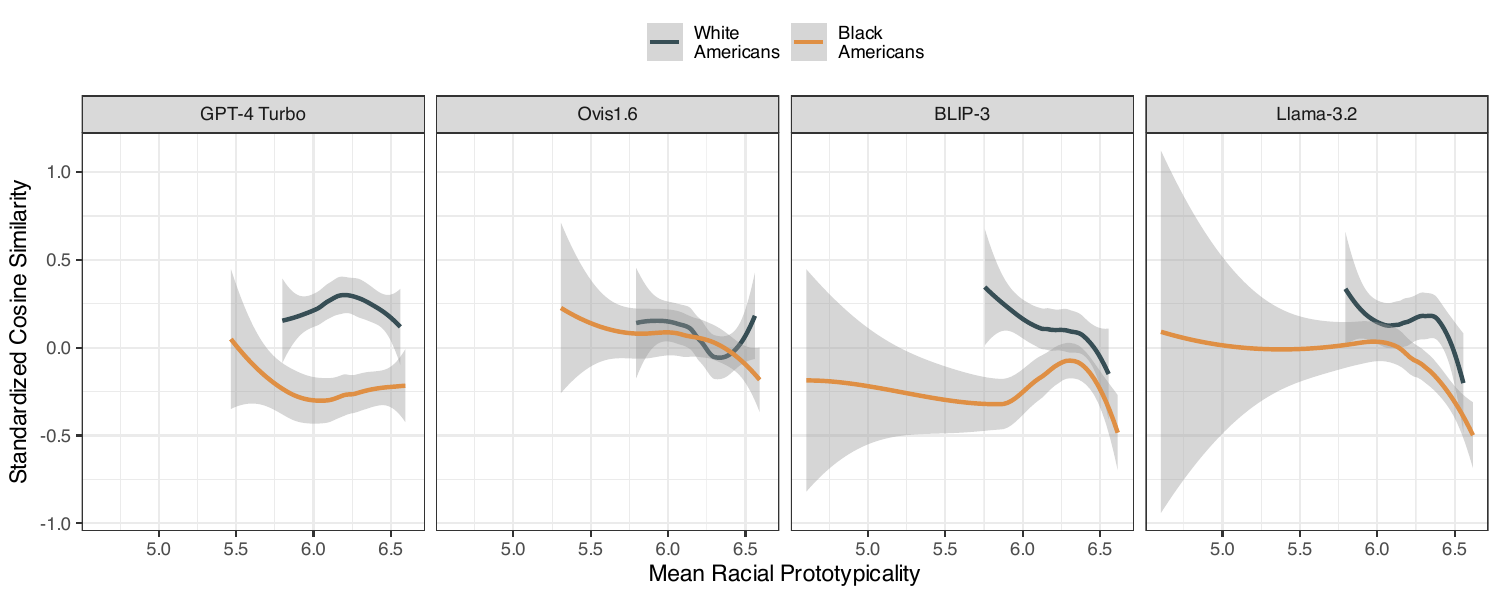}
    \end{subfigure}
    \caption{Standardized cosine similarity (10,000 random samples for each racial group) by prototypicality, calculated using all three encoder models. The top and bottom 10\% of prototypicality values were excluded to minimize tail effects.}
    \label{Figure: Race Interactions (Supplement)}
\end{figure*}


\begin{table*}[!htbp]
    \caption{Topics identified from the Race STM.}
    \label{Table: STM Topics (Race)}
    \centering
    \footnotesize
    \vspace{1em}
    
    \begin{tabular}{C{0.05\textwidth} C{0.10\textwidth} C{0.12\textwidth} C{0.10\textwidth} m{0.50\textwidth}} 
    \toprule
    \textbf{Topic} & \textbf{Proportions} & \textbf{FREX Words} & \textbf{Topic Label} & \textbf{Example} \\ \midrule
    1 & 8.46\% & communiti, school, local, basketball, share & School & On a typical day, our protagonist rises early, driven by a passion for basketball. After a quick breakfast, he heads to school, where he excels academically and plays on the varsity team. In the afternoons, he practices at the local court, honing his skills. As the sun sets, he enjoys time with friends, discussing future dreams and aspirations. The day concludes with quiet reflection, fueling his commitment to excellence and dreams of a brighter future. \\ \midrule
    2 & 14.77\% & help, gentl, reflect, began, simpl & -- & In the heart of a bustling American city, the day began with a soft morning light. Alex, with his casual yet stylish attire, embraced the day with a quiet resolve. A quick breakfast fueled him for a productive day at his tech startup. Amidst the city's vibrant pulse, he found moments of tranquility, reflecting on the day's challenges and successes. \\ \midrule
    3 & 15.65\% & dinner, good, project, gym, break & -- & As a busy software engineer, Michael often starts his day with a cup of coffee and a quick review of his schedule. He spends his mornings coding and collaborating with his team, solving complex problems and creating innovative solutions. After lunch, he might take a short walk around the office building to clear his head and recharge. In the late afternoon, he enjoys a quick chat with his colleagues before diving back into work. As the day winds down, he wraps up any remaining tasks and prepares for the next day. \\ \midrule
    4 & 19.87\% & jazz, novel, everi, sketch, paint & Art & Each morning, she brews coffee, savoring its aroma. She bikes to her art studio, savoring San Francisco's breeze. Hours pass amidst vivid canvases, painting visions of freedom. At noon, she strolls the golden city's bustling streets, seeking inspiration. Evening arrives with soothing guitar melodies under a starry sky, dreams alight." \\ \midrule
    5 & 8.86\% & alarm, check, yawn, rub, clock & -- & The young man woke up to the sound of his alarm blaring in his ear. He rubbed the sleep from his eyes and swung his legs over the side of the bed. After a quick shower, he grabbed a granola bar and headed out the door to catch the bus to work. \\ \midrule
    6 & 9.17\% & patient, care, lab, hospit, nurs & Healthcare & Kelly wakes up early to get ready for her day at work. She brushes her teeth, takes a shower, and gets dressed in her work clothes. Kelly works as a nurse in a busy hospital, so she knows that her day will be busy and challenging. She makes sure to stay focused and take care of her patients to the best of her ability. After her shift, Kelly goes home to rest and prepare for her night shift. She loves her job and the people she works with, and she is always grateful for the opportunity to help others. \\ \midrule
    7 & 16.74\% & thought, lost, gaze, rose, sip & -- & As he sipped his morning coffee, he gazed out the window, lost in thought. The sun rose higher, casting a warm glow over the city. He took a deep breath, feeling the weight of the day ahead. With a quiet determination, he rose from his chair, ready to face whatever challenges lay in store. Today would be a good day. \\ \midrule
    8 & 6.48\% & hair, shirt, cur, woman, beard & Appearance & A young woman with short dark hair and a shaved headline. She's wearing a white shirt and has a serious expression on her face. She's standing in a white background, staring off into the distance. She's not smiling, but her expression is calm and composed. \\ \bottomrule
    \end{tabular}
\end{table*}

\begin{table*}[!htbp]
    \caption{The effect of race and the interaction between race and racial prototypicality in individual VLMs across three topics. A significant positive race effect indicates that the topic is more prevalent for Black Americans than White Americans. A significant interaction term suggests that prototypicality has a stronger influence on topic prevalence for Black Americans compared to White Americans. Significant effects are marked with * when the 95\% confidence intervals do not overlap with 0.}
    \label{Table: Trait Associations (Race)}
    \centering
    \footnotesize
    \vspace{1em}
    
    \begin{tabular}{l cccc}
    
        \toprule
        & \multicolumn{4}{c}{\textbf{Race}} \\ \cmidrule{2-5}
        & \textbf{Basketball} & \textbf{Art} & \textbf{Healthcare} & \textbf{Appearance} \\ \midrule 
        GPT-4 Turbo & 0.051$^{*}$ & -0.020$^{*}$ & 0.0015 & 0.00031 \\
        & [0.040, 0.063] & [-0.033, -0.0074] & [-0.010, 0.013] & [-0.0097, 0.010] \\ [1ex]
        Ovis1.6 & 0.14$^{*}$ & -0.0073 & -0.053$^{*}$ & -0.0021 \\
        & [0.12, 0.16] & [-0.028, 0.013] & [-0.074, -0.032] & [-0.020, 0.016] \\ [1ex]
        BLIP-3 & 0.027$^{*}$ & -0.00087 & 0.0030 & 0.057$^{*}$ \\
        & [0.0066, 0.047] & [-0.021, 0.019] & [-0.023, 0.029] & [0.037, 0.078] \\ [1ex]
        Llama-3.2 & 0.052$^{*}$ & 0.0023 & 0.012 & 0.00028 \\
        & [0.032,0.072] & [-0.018, 0.022] & [-0.0080, 0.033] & [-0.017, 0.018] \\ \midrule 
    
        \toprule
        & \multicolumn{4}{c}{\textbf{Interaction}} \\ \cmidrule{2-5}
        & \textbf{Basketball} & \textbf{Art} & \textbf{Healthcare} & \textbf{Appearance} \\ \midrule 
        GPT-4 Turbo & 0.023$^{*}$ & -0.031 & 0.0027 & 0.0064 \\
        & [0.0029, 0.043] & [-0.063, 0.000036] & [-0.019, 0.024] & [-0.012, 0.025] \\ [1ex]
        Ovis1.6 & 0.047$^{*}$ & 0.0050 & -0.030 & 0.0071 \\
        & [0.0082, 0.085] & [-0.042, 0.052] & [-0.068, 0.0085] & [-0.026, 0.040] \\ [1ex]
        BLIP-3 & 0.0061 & 0.0020 & 0.019 & -0.0088 \\
        & [-0.028, 0.041] & [-0.045, 0.049] & [-0.030, 0.067] & [-0.045, 0.028] \\ [1ex]
        Llama-3.2 & 0.020 & 0.0058 & 0.0051 & 0.0012 \\
        & [-0.015, 0.055] & [-0.041, 0.052] & [-0.032, 0.043] & [-0.032, 0.034] \\ \bottomrule 
    
    \end{tabular}
\end{table*}

\clearpage
\newpage

\begin{figure*}[!htbp]
    \centering
    \includegraphics[width=\textwidth]{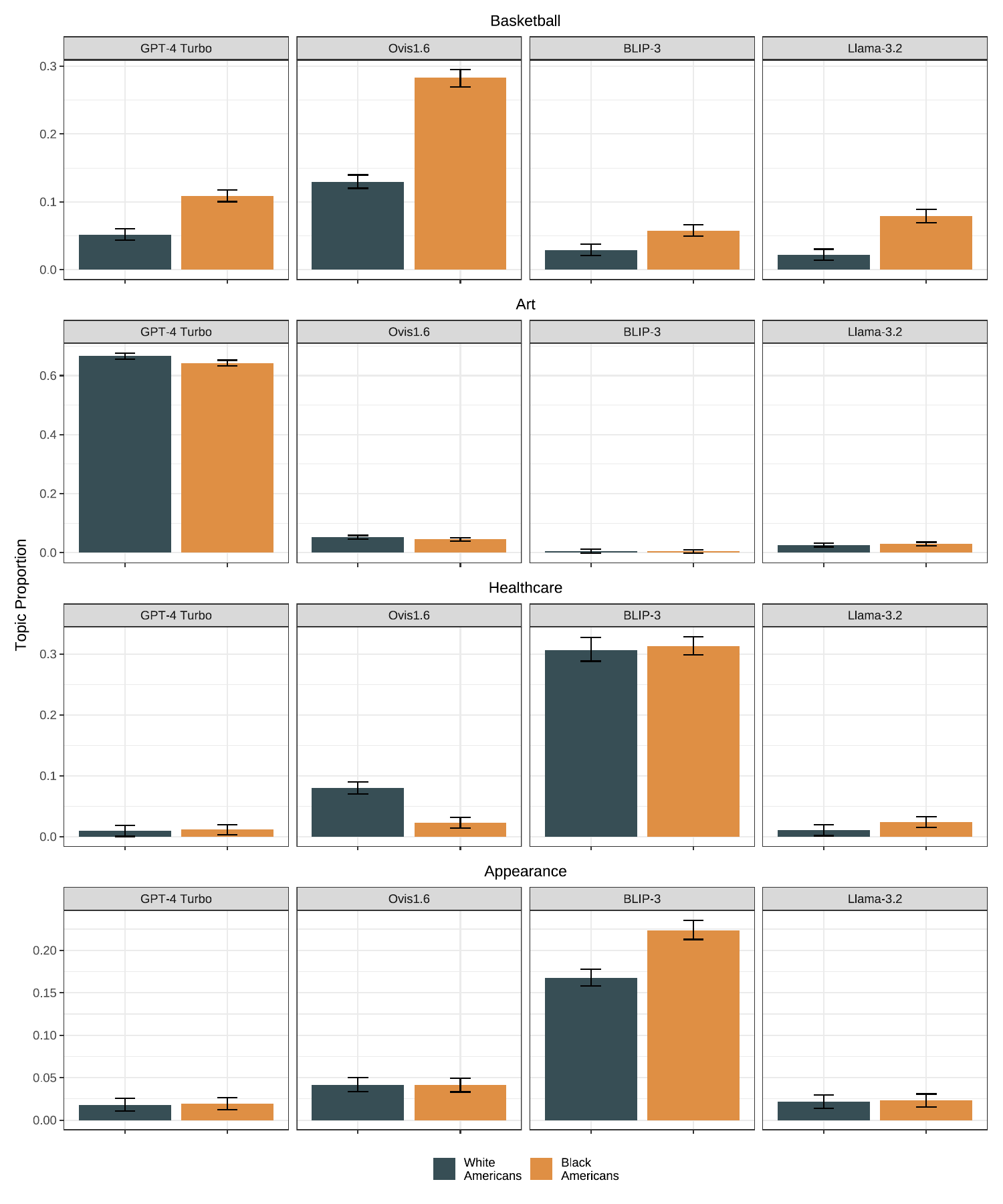}
    \caption{Prevalence of all four topics across four VLMs. Error bars indicate 95\% confidence intervals.}
    \label{Figure: Trait Associations (Race)}
\end{figure*}

\end{document}